%% file: main.tex
\documentclass[pdflatex,sn-mathphys-num]{sn-jnl}% Math and Physical Sciences Numbered Reference Style
\usepackage{graphicx}%
\usepackage{subcaption}
\usepackage{multirow}%
\usepackage{amsmath,amssymb,amsfonts}%
\usepackage{amsthm}%
\usepackage{mathrsfs}%
\usepackage[title]{appendix}%
\usepackage{xcolor}%
\usepackage{textcomp}%
\usepackage{manyfoot}%
\usepackage{booktabs}%
\usepackage{algorithm}%
\usepackage{algorithmicx}%
\usepackage{algpseudocode}%
\usepackage{listings}%
\usepackage{array}
\usepackage{pdflscape}
\usepackage{graphicx}
\usepackage{amssymb}
\usepackage{pifont}
\usepackage{todonotes}

\usepackage{tikz, xcolor, pifont}
\usepackage{graphicx}
\usepackage{caption}
\usepackage{pgfplots}

\definecolor{clr1}{RGB}{255,215,0}
\definecolor{clr2}{RGB}{255,177,78}
\definecolor{clr3}{RGB}{250,135,117}
\definecolor{clr4}{RGB}{234,95,148}
\definecolor{clr5}{RGB}{205,52,181}
\definecolor{clr6}{RGB}{157,2,215}
\definecolor{clr7}{RGB}{0,0,255}

\definecolor{green1}{RGB}{153, 204, 153} 
\definecolor{green2}{RGB}{120, 190, 120} 
\definecolor{green3}{RGB}{180, 220, 180} 

\definecolor{blue1}{RGB}{173, 216, 230} 
\definecolor{blue2}{RGB}{189, 222, 245}

\definecolor{red1}{RGB}{255, 182, 193}
\definecolor{red2}{RGB}{255, 160, 160}
\definecolor{red3}{RGB}{255, 204, 203}

\definecolor{orange1}{RGB}{255, 204, 153}
\definecolor{orange2}{RGB}{255, 178, 102}
\definecolor{orange3}{RGB}{255, 229, 180}

\definecolor{purple1}{RGB}{216, 191, 216}
\definecolor{purple2}{RGB}{221, 160, 221}
\definecolor{purple3}{RGB}{230, 210, 250}

\definecolor{yellow1}{RGB}{255, 255, 153}
\definecolor{yellow2}{RGB}{255, 255, 204}

\definecolor{teal1}{RGB}{180, 240, 230} 
\definecolor{teal2}{RGB}{120, 200, 190} 
\definecolor{teal4}{RGB}{80, 160, 150}  
\definecolor{teal3}{RGB}{0, 130, 120}  

\usepackage{soul}
\usepackage{xcolor}
\usepackage[table]{xcolor}
\definecolor{lightgreen}{RGB}{204, 230, 204}
\definecolor{verylightgreen}{RGB}{230, 245, 230}
\definecolor{mediumgreen}{RGB}{153, 204, 153}
\definecolor{lightblue}{RGB}{173, 216, 230}   
\definecolor{lightgray}{RGB}{211, 211, 211}   
\definecolor{softpeach}{RGB}{255, 218, 185}  
\definecolor{pastellavender}{RGB}{230, 215, 230}

\definecolor{spatial}{HTML}{A8DADC}      % light blue
\definecolor{comparative}{HTML}{F4A261}  % light orange
\definecolor{statistical}{HTML}{E76F51}  % light red
\definecolor{distribution}{HTML}{2A9D8F} % teal/green

\usepackage{makecell}

\theoremstyle{thmstyleone}%
\theoremstyle{thmstyletwo}%

\theoremstyle{thmstylethree}%

\begin{document}

\title[Exploring SAIG Methods for an Objective Evaluation of XAI]{Exploring SAIG Methods for an Objective Evaluation of XAI}
% A Review of SAIG Methods to Evaluate XAI Objectively
% Objective Evaluation of XAI: A Review of SAIG Methods
% Exploring SAIG Methods for an Objective Evaluation of XAI


\author[1,2]{\fnm{Miquel} \sur{Miró-Nicolau}}\email{miquel.miro@uib.cat}
\equalcont{These authors contributed equally to this work.}

\author[1,2]{\fnm{Gabriel} \sur{Moyà-Alcover}}\email{gabriel.moya@uib.cat}
\equalcont{These authors contributed equally to this work.}

\author*[3]{\fnm{Anna} \sur{Arias-Duart}}\email{anna.ariasduart@bsc.es}
\equalcont{These authors contributed equally to this work.}

\affil[1]{\orgdiv{UGiVIA Research Group}, \orgname{University of the Balearic Islands, Dpt. of Mathematics and Computer Science}, \orgaddress{\city{Palma}, \postcode{07122}, \state{Balearic Islands}, \country{Spain}}}

\affil[2]{\orgdiv{Laboratory for Artificial Intelligence Applications (LAIA@UIB)}, \orgname{University of the Balearic Islands, Dpt. of Mathematics and Computer Science}, \orgaddress{\city{Palma}, \postcode{07122}, \state{Balearic Islands}, \country{Spain}}}

\affil[3]{\orgname{Barcelona Supercomputing Center (BSC)}, \orgaddress{\city{Barcelona}, \postcode{08034}, \state{Catalonia}, \country{Spain}}}


\abstract{
The evaluation of eXplainable Artificial Intelligence (XAI) methods is a rapidly growing field, characterized by a wide variety of approaches. This diversity highlights the complexity of the XAI evaluation, which, unlike traditional AI assessment, lacks a universally correct ground truth for the explanation, making objective evaluation challenging. One promising direction to address this issue involves the use of what we term Synthetic Artificial Intelligence Ground truth (SAIG) methods, which generate artificial ground truths to enable the direct evaluation of XAI techniques. This paper presents the first review and analysis of SAIG methods. We introduce a novel taxonomy to classify these approaches, identifying seven key features that distinguish different SAIG methods. Our comparative study reveals a concerning lack of consensus on the most effective XAI evaluation techniques, underscoring the need for further research and standardization in this area.}




\maketitle

\section{Introduction}\label{sec1}

Artificial Intelligence (AI) has become ubiquitous in modern society, with applications in numerous fields obtaining impressive results. Much of this progress began with advances in machine learning and, in particular, with the development of the first functional deep learning models~\cite{krizhevsky2012imagenet}.

In computer vision, deep learning, especially through Convolutional Neural Networks (CNNs), has enabled significant advances in tasks such as image classification, object detection, and segmentation. More recently, the introduction of transformer-based architectures and large-scale multimodal pretraining has led to the emergence of Vision-Language Models (VLMs), which are capable of jointly processing and reasoning over visual and textual modalities. Although recent VLMs promise powerful capabilities, they still face critical limitations: they struggle with negation~\cite{alhamoud2025vision}, basic geometric reasoning~\cite{rahmanzadehgervi2024vision}, data leakage~\cite{chenwe}, and reliability in sensitive tasks such as medical diagnosis~\cite{yanworse}. Combined with their high computational costs, these shortcomings limit their applicability in many real-world settings.   

As a result, simpler architectures like CNNs remain widely used due to their efficiency, scalability and strong performance in many vision tasks. Yet, like larger models, CNNs also suffer from a lack of interpretability, a challenge commonly referred to as the ``black-box problem''~\cite{barredoarrieta2020explainable}. As Vilone and Longo~\cite{vilone2021notionsa} note, the term ``black-box'' refers to the fact that most deep learning models are ``extremely difficult to interpret and explain to laypeople''.

%Deep learning models are characterized by their high complexity compared to more traditional, shallow approaches. However, this complexity introduces a significant challenge: the so-called ``black-box problem''~\cite{barredoarrieta2020explainable}. As noted by Vilone and Longo~\cite{vilone2021notionsa}, the term ``black-box'' refers to the fact that most deep learning models are ``extremely difficult to be interpreted and explained to laypeople''.

To address this issue, the field of eXplainable Artificial Intelligence (XAI) has emerged. XAI seeks to shift the focus towards more transparent AI systems by developing a suite of techniques that improve interpretability while preserving high levels of performance~\cite{adadi2018peeking}. In recent years, XAI has become a highly active field, encompassing various lines of research~\cite{longo2024explainable}. XAI methods can be broadly categorized into two main approaches: \textit{ante-hoc} and \textit{post-hoc} explainability.

\textit{Ante-hoc} approaches are characterized by the use of inherently transparent models, such as linear regression, decision trees, and k-nearest neighbors. While these models offer interpretability by design, a major limitation lies in their comparatively lower performance when contrasted with black-box models. Moreover, increasing the complexity of these otherwise simple models can also render them opaque~\cite{speith2022review}. To overcome these limitations, recent works have proposed complex but explainable-by-design models, such as networks like ProtoPNet~\cite{chen2019looks} and PIP-Net~\cite{nauta2023pip}, that aim to combine interpretability with competitive accuracy.

On the other hand, \textit{post-hoc} methods aim to explain already trained black-box models while preserving their predictive performance. This category is the most widely used in practice, which has led to the development of a wide variety of \textit{post-hoc} techniques in the literature~\cite{zeiler2014visualizing, simonyan2014deep, bach2015pixelwise, ribeiro2016why, zhou2016learning, chattopadhay2018gradcam}. However, this diversity has led to the so-called disagreement problem. First defined by Krishna \emph{et al.}~\cite{krishna2024disagreement}, the disagreement problem is the fact that ``explanations generated by various methods disagree with each other – \textit{e.g.,} the top-k most important features output by different methods may differ''.

To address the disagreement problem, different approaches have been proposed to evaluate XAI methods. According to multiple authors~\cite{nauta2023anecdotal, amengual-alcover2025evaluation, vilone2021notionsa, bodria2023benchmarking, doshi-velez2018considerations}, evaluation mainly falls into two categories: human based analysis (also known as human-grounded~\cite{doshi-velez2018considerations, bodria2023benchmarking}, and plausability~\cite{nauta2023anecdotal}) and machine centered analysis (also known as functionally grounded~\cite{doshi-velez2018considerations, bodria2023benchmarking}, correctness~\cite{nauta2023anecdotal}, and objective evaluation~\cite{vilone2021notionsa}). Human based analysis contains ``those studies that evaluated methods for explainability with a human-in-the-loop approach by involving end-users''~\cite{vilone2021notionsa}. Biecek and Samek~\cite{biecek2024position} identify these two approaches as two different evaluations cultures: RED (for machine centred approaches) and BLUE (for human centered approaches).

While the importance of human-centered evaluation is undeniable, it relies on the assumption that the explanations are faithful to the underlying model; if this faithfulness is lacking, human evaluation alone can be misleading. Therefore, in this work, we focus on machine-centered evaluations, as we believe that building truly trustworthy XAI systems requires prioritizing the assessment of alignment between explanations and model behavior.

Since there is no GT for what constitutes a correct explanation, some methods address this challenge by creating synthetic datasets or methodologies that include GT (or semi GT) explanations. In this work, we introduce the term Synthetic Artificial Intelligence Ground truth (SAIG) to refer to these methods. Our study specifically focuses on SAIG methods that leverage these ground-truth explanations to objectively evaluate XAI techniques.

%Particularly, we aimed to analyse the methods that aimed to surpass the limitations of XAI metrics. Therefore, our study is centred on the RED culture of evaluation techniques~\cite{biecek2024position}. 
This paper has two main contributions: (1) a review of the SAIG state-of-the-art approaches to objectively evaluate XAI methods, and (2) the proposal of a taxonomy to classify and compare these methods. The taxonomy is designed to highlight similarities between approaches and uncover current trends in the field. More broadly, our objective is to advance the evaluation landscape by introducing a systematic perspective that is currently lacking in much of the existing literature.

The document is organized as follows. Section~\ref{sec:machine-centered} introduces the various types of machine-centered explanations. Section \ref{sec:saig-methods} reviews the image-based SAIG methods identified in the literature. Section~\ref{sec:xai_methods} describes the XAI methods that are evaluated using SAIG approaches. Section~\ref{sec:taxonomy} introduces our proposed taxonomy and classifies the reviewed methodologies accordingly. Section~\ref{sec:discussion} analyzes patterns and relationships among SAIG approaches. Finally, Section~\ref{sec:conclusions} concludes the work and discusses potential future directions for the field.

%\todo[inline]{Afegir un paragraf que expliqui l'estructura del paper}

% Two main approaches have been proposed for machine-centred evaluation. The first, known a XAI metrics, focuses on assessing specific properties of explanations based on assumed relationships between the model’s input, output, and the generated explanation.  Within this category we find fidelity metrics~\cite{bach2015pixelwise, samek2017evaluating, alvarez-melis2018robustness}, robustness metrics~\cite{alvarez-melis2018robustness, agarwal2022rethinking, yeh2019infidelity}, among others. The main advantage of these approaches is their applicability in real-world scenarios. 

\section{XAI evaluation} \label{sec:machine-centered}

Machine-centered evaluation, those that ``employed objective metrics and automated approaches to evaluate methods for explainability''~\cite{vilone2021notionsa}, typically rely on XAI metrics. These approaches focuses on assessing specific properties of explanations based on assumed relationships between the model’s input, output, and the generated explanation.  Within this category we find fidelity metrics~\cite{bach2015pixelwise, samek2017evaluating, alvarez-melis2018robustness}, robustness metrics~\cite{alvarez-melis2018robustness, agarwal2022rethinking, yeh2019infidelity}, among others. However, several authors have raised criticisms regarding their reliability~\cite{tomsett2020sanity, hedstrom2023metaevaluation, miro-nicolau2024metaevaluating, miro-nicolau2025comprehensive}. In particular, for image data, the generation of out-of-distribution samples can lead models to behave unpredictably, rendering the resulting metrics unreliable and limiting the validity of any conclusions drawn from them~\cite{miro-nicolau2025comprehensive}. This challenge highlights the need for more robust evaluation strategies in the vision domain and motivates our focus on image-based XAI evaluation in this work.

%In the introduction we have noted the importance of XAI evaluation. However, XAI metrics, the main machine centred approach to measure the quality of explanation, has been largely critisized, with multiple authors identifying a set of worrisome limitations~\cite{tomsett2020sanity, hedstrom2023metaevaluation, miro-nicolau2024metaevaluating, miro-nicolau2025comprehensive}. 

The complexity of the previous metrics is what Hëdstrom \emph{et al.}~\cite{hedstrom2023metaevaluation} called the \textit{challenge of unveriafibility}. These authors noted that the limits of the evaluation arise from the lack of an available ground truth (GT) for the explanation. Multiple explanation can be correct to solve a task, but for a singular element and a particular element only exist one correct explanation. % This fact is known in the state-of-the-art as the Rashomon effect~\cite{rudin2019stop}.

Aiminig to solve the challenge of \textit{unveriafibility} a field of study emerged around the development of synthetic datasets that include built-in ground-truth (or semi ground-truth) explanations. As introduced earlier, we refer to this paradigm as Synthetic Artificial Intelligence Ground truth (SAIG). Several works in the literature have proposed SAIG datasets~\cite{agarwal2022rethinking, cortez2013using, yang2019benchmarking, guidotti2021evaluating, arras2022clevr, arias2022focus, miro2024assessing, bastings2022will}. Although this paper focuses exclusively on image data \cite{arras2022clevr, arias2022focus, miro2024assessing, guidotti2021evaluating}, the field spans multiple data modalities, including tabular~\cite{cortez2013using, guidotti2021evaluating, agarwal2022openxai}, and text~\cite{bastings2022will}.

Despite the increasing interest in SAIGs datasets, the field still lacks a unified terminology, a shared taxonomy, and a thorough analysis of prior work. To the best of our knowledge, there is currently no comprehensive review or taxonomy specifically focused on SAIG approaches in the XAI literature. As new SAIG methodologies emerge all these limitations suposse a major setback for the field’s development.

%\todo[inline]{Explicar perquè feim imatges.}

\section{SAIG Methods} \label{sec:saig-methods}

We identified sixteen different SAIG image proposals, as summarized in Table \ref{tab:methods-table}. These studies exhibit two notable similarities: most do not analyze prior state-of-the-art approaches and can be categorized based on a shared set of characteristics.

\paragraph{\href{https://doi.org/10.24963/ijcai.2017/371}{Right for the Right Reasons}}

Although explanation evaluation is not the main goal of this work \cite{ross2017right}, Ross \emph{et al.} introduce two SAIG datasets designed to evaluate model explanations. The \textit{Toy Color} dataset consists of simple 5×5×3 RGB images with class labels based on specific color patterns in the corners and top-middle pixels, these are the GT features for explanation. The \textit{Decoy MNIST} dataset modifies standard MNIST by adding gray switches in the corners whose shades depend on the label during training but become random at test time, thus, the digit shape is the GT feature. %In both datasets, the relevant explanatory features (\textit{i.e., the GT}), are implicitly defined by how the data is constructed.

\paragraph{\href{https://openreview.net/forum?id=H1ziPjC5Fm}{Visual Explanation by Interpretation}}

Oramas \emph{et al.} \cite{oramasvisual} introduced the \textit{an8Flower} dataset, a synthetic dataset designed for evaluating explanations. It consists of flower images where different parts of the object, such as petals or stems, change color to define multiple classes. For example, the an8Flower-single-6c variant contains 6 classes distinguished by color changes in a single part. In the an8Flower-double-12c variant, both the color and the specific part that changes color serve as discriminative features defining the 12 classes. The GT explanations are provided as masks highlighting these parts specific to each class.

\paragraph{\href{https://arxiv.org/pdf/1907.09701}{Benchmarking Attribution Methods with Relative Feature Importance}}

Yang \emph{et al.} \cite{yang2019benchmarking} proposed the \textit{Benchmarking Attribution Methods (BAM)}  dataset, a synthetic dataset in which GT feature relevance is controlled. This control is achieved by pasting objects from the MSCOCO \cite{lin2014microsoft} dataset into scene images from the MiniPlaces \cite{zhou2017places} dataset. Each composite image has both an object label (\textit{e.g.,} dog) and a scene label (\textit{e.g.,} bedroom), allowing models to be trained to predict either the object (relying on the pasted region) or the scene (relying on the background). This setup defines which regions of the image should be relevant to the prediction and which should not.

\paragraph{\href{https://proceedings.neurips.cc/paper_files/paper/2021/file/0fe6a94848e5c68a54010b61b3e94b0e-Paper.pdf}{Do Input Gradients Highlight
Discriminative Features?}}

Shah \emph{et al.} \cite{shah2021input} introduced the \textit{BlockMNIST} dataset, a synthetic benchmark based on MNIST. Each image is constructed by vertically concatenating two 28×28 image blocks: a signal block (an MNIST digit image of 0 or 1) and a null block containing non-discriminative patterns. The signal and null blocks are randomly placed at the top or bottom of the image with equal probability. Only the signal block determines the class label (\textit{i.e.,} the GT for feature relevance), while the null block is uninformative.

\paragraph{\href{https://doi.org/10.1016/j.artint.2020.103428}{Evaluating local explanation methods on ground truth}}

Guidotti \emph{et al.} \cite{guidotti2021evaluating} developed the 
\textsc{seneca-img}, a generator of synthetic transparent image classifiers from which it is possible to extract synthetic GT explanations. The authors evaluate explanation methods using this transparent classifier, which uses a simple pattern matching approach, along with a dataset consisting of synthetic images of fixed size with a black background. Each image is divided into a grid of smaller cells, and a pattern is inserted into half of the images to train and test the classifier. Since the classifier’s decision depends solely on the presence of this pattern, the pattern itself serves as the GT explanation.

%The synthetic classifier produced by \textsc{seneca-img} employs a simple pattern matching approach, rather than a neural network, to scan the image, detect the presence or absence of the pattern, and perform classification accordingly. It also outputs a GT explanation where the pixels corresponding to the pattern are marked as relevant, and the rest as irrelevant. This GT explanation allows the quantitative evaluation of explanation methods.

\paragraph{\href{https://doi.org/10.1109/FUZZ-IEEE55066.2022.9882821}{Focus! Rating XAI Methods and Finding Biases}}

Arias-Duart \emph{et al.}~\cite{arias2022focus} introduced the 
\textit{Focus}, an evaluation score for feature attribution methods that uses mosaics as GT. These mosaics are created by combining images from different classes within the original data distribution. \textit{Focus} measures the proportion of relevance that falls within the regions corresponding to the relevant classes. Since \textit{Focus} functions similarly to a precision metric, other classification metrics such as recall and F1-score can also be applied to evaluate explanation methods alongside mosaics, as discussed in \cite{arias2023confusion}. Importantly, the mosaic methodology can be applied to any dataset without requiring model retraining.

\paragraph{\href{https://doi.org/10.1016/j.inffus.2021.11.008}{CLEVR-XAI}}

Arras \emph{et al.}~\cite{arras2022clevr} proposed a methodology for generating a SAIG based on visual question answering (VQA) models. These models are specific types of AI models that can answer questions regarding the content of an image. These authors proposed to define a set of well crafted questions, allowing to know whether and object is important or not. 

\paragraph{\href{https://doi.org/10.1175/AIES-D-22-0012.1}{Investigating the fidelity of explainable artificial intelligence methods}}

Mamalakis \emph{et al.}~\cite{mamalakis2022investigating} defined a set of images containing circular and square frames of random size. They proposed a binary classification task: the first class corresponds to instances where the area of the circle is larger than that of the square, and the second class to the opposite case. Consequently, the authors used the expected sign of the explanation as the GT: the first class the pixels corresponding to circular objects must have positive attributions, while the square ones negatives.

\paragraph{\href{https://doi.org/10.1109/CVPR52688.2022.00998}{Towards Better Understanding Attribution Methods}}

Rao \emph{et al.}~\cite{rao2022better} built upon previous mosaic-based approaches~\cite{arias2022focus, bohle2021convolutional}. Their main contribution is the use of modified models that ensure the predictions for different classification classes are not influenced by one another.

\paragraph{\href{https://proceedings.mlr.press/v162/kim22h.html}{Sanity Simulations for Saliency Methods}}

Kim \emph{et al.}~\cite{kim2022sanity} proposed to combine a set of synthethic shapes (letters and squares) in a uniform background to generate the image SAIG. The authors trained a binary classification model, the classes depend on the presence or not of some of the synthethic shapes. This setup allows the authors to know which elements are important and which are not.

\paragraph{\href{https://doi.org/10.1109/ICCV51070.2023.00368}{FunnyBirds}}

Hesse \emph{et al.}~\cite{hesse2023funnybirds} proposed a synthetic vision dataset consisting of 3D birds designed using five human-comprehensible concepts: beak, wings, feet, eyes, and tail. Each FunnyBird class consists of a unique combination of these parts and a bird body. This configuration allows for evaluating the importance of the different parts without having to predefine it beforehand.

\paragraph{\href{https://doi.org/10.1016/j.artint.2024.104179}{Assessing fidelity in XAI post-hoc techniques}}

Miró-Nicolau \emph{et al.}~\cite{miro2024assessing} proposed a SAIG composed of simple geometric shapes. They adapted a previous approach based on tabular datasets, enabling the generation of an XAI GT with predefined and specific importance values for each pixel. Finally, they compared the saliency maps produced by 16 XAI methods to this GT, treating both as probability distributions.

\paragraph{\href{https://openaccess.thecvf.com/content/CVPR2021/html/Bohle_Convolutional_Dynamic_Alignment_Networks_for_Interpretable_Classifications_CVPR_2021_paper.html}{Convolutional Dynamic Alignment Networks}}

Böhle \emph{et al.}~\cite{bohle2021convolutional} introduced Dynamic Alignment Units (DAUs), which improve the interpretability of neural networks. To evaluate them, they used two strategies: first, they assessed a localization metric by adapting the pointing game to the CIFAR-10 and TinyImageNet datasets, evaluating attribution methods on a grid of n×nn×n images; second, they analyzed the model's behavior using a pixel removal strategy.

\paragraph{\href{https://doi.org/10.1145/3630106.3658537}{Classification Metrics for Image Explanations}}

Fresz \emph{et al.}~\cite{fresz2024classification} presented an extension of the work by Arias-Duart \emph{et al.}~\cite{arias2022focus}, considering negative Feature Importance (FI). Both positive and negative FI are taken into account for the specification of true positives and negatives, as well as false positives and negatives, with respect to the FI on the mosaics. A complete confusion matrix can thus be defined, enabling the computation of metrics commonly applied in classification tasks.

\paragraph{\href{https://doi.org/10.1609/aaai.v36i9.21196}{Do Feature Attribution Methods Correctly Attribute Features?}}

Zhou \emph{et al.}~\cite{zhou2022feature} proposed evaluating attribution methods datasets systematically modified to introduce ground truth information for attributions, they called natural datasets. The modifications (blurring, brightness change, hue shift, pixel noise and watermark) ensures that any classifier with sufficiently high performance has to rely, sometimes solely, on the manipulations. They studied the attribution percentage assigned to the joint effective region as their main metric.

\paragraph{\href{https://doi.org/10.1109/TAI.2022.3228834}{Quantifying Explainability of Saliency Methods}}

Tjoa \emph{et al.}~\cite{tjoa2022quantifying} created a synthetic dataset with in-built GT heatmaps. They generated unambiguous discrete-valued attribution heatmaps as explanation and introduced of metric that computes pixelwise hits and misses. The dataset is also customizable and augmented with noisy variations of basic shapes.  

\section{XAI methods} \label{sec:xai_methods}

The vast majority of the SAIG discussed in the previous section not only propose approaches for obtaining an XAI ground truth, but also use their proposals to compare multiple state-of-the-art XAI methods. Table~\ref{tab:xai_methods} summarizes the XAI methods analyzed across the different SAIG methods, highlighting in green the best-performing method according to each respective study. In total, 32 distinct XAI methods are evaluated.

The primary distinction between XAI methods was already discussed in the introduction: \textit{ante-hoc} or \textit{post-hoc} methods. Among the 32 methods used in the reviewed SAIG approaches, only four are \textit{ante-hoc}. Moreover, when counting the number of times each method is applied, \textit{ante-hoc} methods are used only five times, compared to 121 uses of \textit{post-hoc} methods. We expected this result: \textit{ante-hoc} are transparent by definition and, therefore, the evaluation (from a machine-centred point of view) of their explainability quality is superfluous. 

Additionally, XAI methods can be categorised according to multiple features. We classify them both depending on their inner working and output type. Barredo-Arrieta \emph{et al.}~\cite{barredoarrieta2020explainable} define six categories depending on the output type. Within the SAIG papers we reviewed we only find two categories from Barredo-Arrieta~\emph{et al.}~\cite{barredoarrieta2020explainable} proposal: \textit{explanation by example} and \textit{feature relevance}. Furthemore, we also consider \textit{explanation by concept}, as a different output type:

\begin{itemize}
    \item \textbf{Feature relevance}. According to Barredo-Arrieta \emph{et al.}~\cite{barredoarrieta2020explainable} ``clarify the inner functioning of a model by computing a relevance score for its managed variables''. Within this category we can find saliency maps in the image context. Examples of these methods are, among others: LRP~\cite{bach2015pixelwise}, Gradient~\cite{simonyan2014deep}, GradCAM~\cite{selvaraju2020gradcam}.
    \item \textbf{Explanations by example}. The same authors \cite{barredoarrieta2020explainable} defined it as methods that ``consider the extraction of data examples that relate to the result generated by a certain model, enabling to get a better understanding of the model itself''. Within this category we identified ProtoPNet~\cite{chen2019looks}. 
    \item \textbf{Explanation by concept}. This category consider all the methods that aimed to explain the prediction according to the importance of some human concept. We identified TCAV~\cite{kim2018interpretability} and BagNet~\cite{brendel2018approximating} as Concept explanations.
\end{itemize}

Miró-Nicolau \emph{et al.}~\cite{miro2024assessing} propose a taxonomy according to the inner working for \textit{post-hoc} feature relevance methods. Following their proposal we categorize into three distincts categories:

\begin{itemize}
    \item \textbf{Backpropgation-based methods}. Set of XAI methods that obtained the explanation propagating the output to the input. Most methods used the built-in backpropagation algorithm of the model, some of them modifying how the non-linearities are handled. Examples of methods from this category are: Smoothgrad~\cite{smilkov2017smoothgrad}, Guided Backpropagation~\cite{springenberg2014striving}, Integrated gradients~\cite{sundararajan2017axiomatic} among others.
    \item \textbf{CAM methods}. XAI methods based on the original work from Zhou \emph{et al.}~\cite{zhou2016learning}. The methods within this category are characterised by obtaining the explanation through the output of a convolutional layer, therefore only being able to use it with CNN. This output, a set of small matrixes, is then upsampled to the size of the image and weighted with some importance puntuaction. Examples of methods from this category are: GradCAM~\cite{selvaraju2020gradcam}, GradCAM++~\cite{chattopadhay2018gradcam} or ScoreCAM~\cite{wang2020scorecam}, among others.
    \item \textbf{Perturbation-based methods}. XAI methods consisting on modifying the input data and analysing the behaviour of the output as a proxy to obtain the explanation. In images the perturbation procedure consists, usually, on the substitution of the orginal values of a part of the pixels for an uniform value. Examples of this approach are well-known methods such as: LIME~\cite{ribeiro2016why}, Occlusion~\cite{zeiler2014visualizing}, among others.
\end{itemize}

From the study of the XAI methods used we can see that SAIG methodologies are used with a large and diverse set of explanaibility techniques. Additionally, from this analysis we can confirm that the XAI methods usage trends, already studied in different reviews of the XAI field~\cite{barredoarrieta2020explainable, longo2024explainable}. Particularly, the predominance of \textit{post-hoc} approaches and the prevelance of saliency maps for images.

\section{Taxonomy of SAIG Methods} \label{sec:taxonomy}

In this section, we detail the common features and propose a taxonomy to organize the SAIG methods accordingly. The taxonomy is organized around five dimensions that characterize SAIG evaluation methods. First, we examine the image features used to construct datasets, including the type of visual elements, their position, and background (\S\ref{sec:image-features}). We then explore how the ground truth is defined (\S\ref{sec:gt-definition}) and encoded  (\S\ref{sec:gt-value}). Next, we assess the generability of these methods, referring to their adaptability to different use cases (\S\ref{sec:generability}). Finally, we categorize the different evaluation metrics used to assess explanation quality, providing a complete view of how explainability methods are benchmarked (\S\ref{sec:eval-metrics}).

\subsection{Image Features} \label{sec:image-features}

This category refers to the visual characteristics that can be precisely controlled to design datasets for evaluating XAI methods. In this section, we examine three features of such images: the type of the evaluation elements, their position within the image, and the background on which they are placed. These features are not independent; choices made in one of them often constrain or determine the others. 

%\hypertarget{type-section}{\subsubsection{\colorbox{green3}{\strut Type}}}

\hypertarget{type-section}{\subsubsection{\strut Type}}

This category refers to the differnt types of visual elements used in evaluation datasets. We identify three main categories (see Figure \ref{fig:type}):

\begin{enumerate}
    \item \textbf{Simple figures.} These typically consist of 2D geometric shapes such as squares, circles, or triangles, \cite{ross2017right, guidotti2021evaluating, mamalakis2022investigating, miro2024assessing,tjoa2022quantifying} and in some cases, textual elements \cite{kim2022sanity}. A few datasets extend this category to include rendered 3D shapes \cite{arras2022clevr}. Simple figures are often associated with uniform backgrounds, allowing for minimal visual interference. This simplicity facilitates the analysis of model behavior and the isolation of specific visual attributes.
    
    \item \textbf{Complex figures.} These are usually composed of multiple polygons or abstract shapes that approximate real-world objects without being photographic in nature. In \cite{hesse2023funnybirds} they constructed a fine-grained bird species dataset inspired by the CUB-200-2011 dataset \cite{WahCUB_200_2011}. In \cite{oramasvisual} they generated a dataset by taking as starting point an eggplant model and they introduced discriminative features to define each of the classes of interest. Finally in \cite{ zhou2022feature} they superimposed visual elements segmented from real-world datasets. In \cite{yang2019benchmarking} the evaluation elements consists on modifications on images.
    
    \item \textbf{Real-world images.} This category includes images sourced from widely used datasets such as ImageNet \cite{russakovsky2015imagenet}, COCO \cite{lin2014microsoft}, CIFAR10 \cite{krizhevsky2009learning}, MIT67 \cite{quattoni2009recognizing}, Places365-Standard \cite{zhou2017places}, Dogs vs. Cats\footnote{\url{https://www.kaggle.com/c/dogs-vs-cats/overview}} and MAMe \cite{pares2022mame}. These images offer high visual complexity and semantic richness. Articles that fall in this category include: \cite{shah2021input, arias2022focus, rao2022better, bohle2021convolutional, fresz2024classification}.
    
\end{enumerate}

\input{type}

%\hypertarget{position-section}{\subsubsection{\colorbox{blue1}{\strut Position}}}

\hypertarget{position-section}{\subsubsection{\strut Position}}

With respect to the spatial positioning of evaluation elements, we identify two main categories (see Figure \ref{fig:positions}):

\begin{enumerate}
    \item \textbf{Random positions.} evaluation elements are randomly assigned across the image space. Articles that fall into this category include \cite{yang2019benchmarking,guidotti2021evaluating, oramasvisual, arras2022clevr,mamalakis2022investigating, kim2022sanity, hesse2023funnybirds, miro2024assessing, zhou2022feature,tjoa2022quantifying}.
    \item \textbf{Predefined positions.} leading to more controlled and interpretable setups.
\end{enumerate}

Among the fixed-position approaches, a common and widely adopted pattern is the use of grid-based arrangements, typically in a $2 \times 2$ layout where evaluation elements are placed in a mosaic. This structure facilitates the definition of position-dependent rules and enables clearer attribution analysis. Most existing evaluation methods fall into this mosaic idea \cite{arias2022focus, fresz2024classification, rao2022better, bohle2021convolutional, shah2021input}.

A different idea of fixed-position approach is introduced in \cite{ross2017right}, which present a dataset composed of $5 \times 5 \times 3$ RGB images with four possible colors. In this case, all evaluation attributes are positioned deterministically. The dataset design embeds two distinct, position-dependent classification rules that a model could implicitly learn: whether the four corner pixels share the same color, and whether the top-middle three pixels each have different colors.

\input{positions.tex}

% COMMENT: Potser estaria be alguna galeria 

\hypertarget{background-section}{\subsubsection{\strut Background}}

%\hypertarget{background-section}{\subsubsection{\colorbox{red3}{\strut Background}}}

We identify three  categories of background configurations based on our findings in the literature (see Figure \ref{fig:background}):

\begin{enumerate}
    \item \textbf{Uniform background.} The entire image background is homogeneous, typically a solid color. This setting minimizes isolates the importance of foreground evaluation elements \cite{guidotti2021evaluating,oramasvisual, arras2022clevr,mamalakis2022investigating, kim2022sanity, miro2024assessing}.
    
    \item \textbf{Compositional background.} Objects or evaluation attributes are superimposed onto background, sampled from natural images or generated textures. This configuration enables the creation of more complex datasets, particularly penalizing occlusion-based XAI methods, as it allows for the generation of out-of-distribution images. 
    
    \item \textbf{No background.} In mosaic or grid-based images where each pixel is part of an evaluation element, the concept of a background becomes irrelevant. In such cases, the entire image area is dedicated to evaluation-relevant content \cite{arias2022focus, fresz2024classification,rao2022better,bohle2021convolutional, shah2021input}.

\end{enumerate}

Regarding compositional background, there are multiple approaches. In \cite{yang2019benchmarking} they constructed a dataset by pasting object pixels from \cite{lin2014microsoft} into scene images from MiniPlaces \cite{zhou2017places}, in \cite{zhou2022feature} they use a self constructed birds dataset as background. In \cite{hesse2023funnybirds} they locate different background objects in the image by using a deterministic algorithm to reflect real-world challenges. In \cite{tjoa2022quantifying} they propose 3 different backgrounds to increase the variation of dataset.

\input{background}

Among all possible combinations of type, position, and background, only six occurrences have been observed, as illustrated in Figure \ref{fig:galeria}.
\input{features}

%\noindent \textcolor{blue}{Proposta galeria imatges amb les combinacions existents}
%\begin{itemize}
%    \item Miquel: Simple, Random, Uniform
%    \item 41 (Ross): Simple, predefined, no background
%    \item Focus!: Real world, predefined, no background
%    \item Funny Birds: Complex, Random, Compositional
%    \item 42 Oramas Complex, random, uniform
%    \item 54 tjoa2022quantifying simple, random, compositional
    
%\end{itemize}

%\hypertarget{gt-definition}{\subsubsection{\colorbox{orange3}{\strut Ground Truth definition}}}\label{sec:gt-definition}
\hypertarget{gt-definition}{\subsection{\strut Ground Truth definition}}\label{sec:gt-definition}

This category refers on how the different author produce the XAI GT. This category and the following, the Ground Truth value, are highly related: how the GT is defined produce a particular type of GT. Therefore, we expected a high dependency between both of them. Based on our findings we identified three main different approaches to generate GT (see Figure \ref{fig:gt_definition}):

\begin{enumerate}
    \item \textbf{Identity}. This approach involves training an AI model to determine whether a specific element is present or absent. Consequently, it is expected that any explanation produced should be localized within that element. This is the most widely used approach, and in fact, the other categories are conceptually derived from it. Within this category, we distinguish between Predefined positions methods \cite{shah2021input, arias2022focus, rao2022better, bohle2021convolutional, fresz2024classification} and Random positions methods \cite{ross2017right, oramasvisual, guidotti2021evaluating, arras2022clevr, kim2022sanity, zhou2016learning, tjoa2022quantifying}, as described in Section \ref{sec:image-features}.
    \item \textbf{\textit{A priori}}. This approach involves training AI model with defined elements importance. The authors of these approaches aimed to defined a SAIG dataset with some kind of relation between the different elements present in the images. We identified three works proposing this: Yang and Kim~\cite{yang2019benchmarking} defined the GT as an important element and none-important element; Mamalakis \emph{et al.}~\cite{mamalakis2022investigating} defined the sign of the relevance of each element, defining an AI output that increase or decrease depending on different elements; Miró-Nicolau \emph{et al.}~\cite{miro2024assessing} defined the AI prediction as following a predefined function that gives different weights to different elements.
    \item \textbf{Interventions}. This approach has been proposed by Hesse \emph{et al.}~\cite{hesse2023funnybirds}. The authors obtained the relevance of each part by removing a particular part from an image and defined its importance as the difference between the AI output after and before this removal operation. As a result obtained local XAI GT, adhered by design to the AI model.  
\end{enumerate}

\input{gt_definition}

\hypertarget{gt-value}{\subsection{\strut Ground Truth Value}}\label{sec:gt-value}

This category refers to the format and values of the generated GT. It defines what is considered a correct explanation and is closely related to the evaluation metrics, which will be discussed later in this paper. We identified three main types of GT values (see Figure \ref{fig:gt_value}):

\begin{enumerate}
    \item \textbf{Binary.} Each pixel in the image is labeled as either important or not important. The resulting GT is, explicitly or implicitly, a binary mask. 
    \item \textbf{Relative.} A relative ranking of image parts is established, where some parts relevance are expected to be more important than others. 
    \item \textbf{Multivalue.} Each pixel is assigned a specific relevance score. variant of this approach considers only the sign of the relevance (positive or negative), while the full version expects an exact numerical value. This results in a saliency map as the GT.
\end{enumerate}

Binary, or positional, GT is the simplest and most intuitive format, and consequently, the most widely used. From the sixteen analysed articles, ten lies within this category \cite{oramasvisual, shah2021input, guidotti2021evaluating, arias2022focus, arras2022clevr, rao2022better, kim2022sanity, bohle2021convolutional, fresz2024classification, zhou2022feature, tjoa2022quantifying}. Only one article proposed a Relative GT value, Yan and Kim~\cite{yang2019benchmarking}. Finally, more recently a set of authors aimed to generate SAIG with specific explanation value \cite{mamalakis2022investigating, hesse2023funnybirds, miro2024assessing}. The increased complexity of these approaches made necessary the usage of more complex measures.

\input{gt_value}

%\hypertarget{portability-section}{\subsubsection{\colorbox{yellow2}{\strut Portability}}}\label{sec:portability}

\hypertarget{generability-section}{\subsection{\strut Generability}}\label{sec:generability}

This category refers to the extent to which an evaluation method can be applied beyond its original context. Some methods are broadly adaptable, while others are tightly coupled to a specific setting and cannot be easily generalized to other use cases. Broadly, we can distinguish between two types of evaluation methods (see Figure \ref{fig:portability}): 

\begin{enumerate}
    \item \textbf{Generalizable methods.} These allow users to build GT data tailored to their specific use cases. Because the evaluation process is built on the user's own dataset, the method remains adaptable. 
    \item \textbf{Non-generalizable methods.} These are based on predefined GT datasets. While they provide insight into the attribution quality, they are limited in scope and cannot be directly applied to new or custom used cases. 
\end{enumerate}

Most existing evaluation methods fall into the non-generalizable category \cite{ross2017right, oramasvisual, yang2019benchmarking, guidotti2021evaluating, arras2022clevr, mamalakis2022investigating, kim2022sanity, hesse2023funnybirds, miro2024assessing, tjoa2022quantifying, zhou2022feature}. However, some approaches \cite{arias2022focus, rao2022better, bohle2021convolutional, fresz2024classification, shah2021input}, particularly those based on mosaic techniques, offer greater generalizability. By allowing users to generate GT using their own dataset images, these methods provide a more flexible and reusable framework for evaluation.

\input{portability}

\hypertarget{evaluation-section}{\subsection{\strut Evaluation Measures}}\label{sec:eval-metrics}

This section categorizes metrics used to evaluate explanation methods into four main types (see Figure \ref{fig:measures}), based on how they assess explanation quality. 

\begin{enumerate}
    \item \textbf{Spatial Alignment Metrics.} These metrics assess how well the explanation aligns spatially with GT regions. Some works, such as \cite{oramasvisual}, use intersection over union (IoU) between thresholded attribution heatmaps and GT masks. However, as noted by \cite{kim2022sanity}, IoU discards valuable information by converting continuous attribution maps into binary masks, ignoring the intensity of the attributions. 

Another approach, used in \cite{arras2022clevr, ross2017right}, measures how many of the top-k most relevant pixels fall within the GT. This metric is referred to as \textit{Relevance Rank Accuracy} in \cite{arras2022clevr}. Similarly, in the BlockMNIST setting \cite{shah2021input}, the fraction of top-k attributions located in the null (non-informative) block is used to quantify incorrect attribution. Although these methods capture whether the most relevant (or least relevant) regions align spatially with GT, they treat all top-ranked pixels equally, ignoring the magnitude of their attribution scores.

To address this, other works preserve the magnitude of the attribution rather than discarding it. These approaches compute the proportion of total attribution that lies within the GT region, similar to precision \cite{arias2022focus, rao2022better, kim2022sanity, bohle2021convolutional, zhou2022feature}. Arias \emph{et al.} \cite{arias2022focus} refer to this as \textit{Focus}, using mosaics for evaluation; Arras \emph{et al.} \cite{arras2022clevr} call it \textit{Relevance Mass Accuracy}; and Zhou \emph{et al.} refer to it as \textit{Attr\%}, where the GT corresponds to the manipulated region.

A related approach is proposed by Hesse \emph{et al.} \cite{hesse2023funnybirds}, where a part is considered important if its cumulative attribution exceeds a percentage of the total. Their synthetic dataset allows defining known sufficient and irrelevant parts, enabling metrics such as the \textit{Controlled Synthetic Data Check (CSDC)}, which measures how well an explanation highlights sufficient parts, and \textit{Distractibility (D)}, which quantifies the extent to which irrelevant parts are mistakenly highlighted.

Finally, some works treat attribution maps as classifier outputs, applying standard classification metrics like precision, recall, F1-score, or ROC curves \cite{fresz2024classification, guidotti2021evaluating, tjoa2022quantifying}. This typically allows evaluations to account for negative relevance as well.

\item \textbf{Statistical Agreement Metrics.} This category assesses the agreement between attribution maps and GT using standard statistical techniques, such as correlation measures \cite{mamalakis2022investigating}. Although these metrics capture the overall value correspondence, they ignore the spatial structure of the attributions.

\item \textbf{Distribution-based metrics.}
This category interprets both attribution maps and GT masks as probability distributions. For example, Miró \emph{et al.} \cite{miro2024assessing} use Earth Mover’s Distance (EMD) \cite{rubner2000earth}, which measures the minimal \textit{work} required to transform one distribution into another, and the Similarity Metric (MIN) \cite{judd2012benchmark}, which quantifies overlap by summing the minimum values at each pixel.

\item \textbf{Comparative Evaluation Metrics.} This category tests whether the attribution methods reflect differences in input or model relevance. The \textit{Model Contrast Score (MSC)} \cite{yang2019benchmarking} compares two models trained on a different setting: one trained on object labels and another on scene labels. The metric measures the difference in attribution for a specific region (\textit{e.g., } a dog), which should be higher in the object-trained model. The \textit{Input Dependence Rate (IDR)}  \cite{yang2019benchmarking} evaluates a single scene trained model by comparing input with and without an object, expecting lower attribution when the object is added. 

\end{enumerate}

\input{evaluation}

\input{method_table}

\section{Discussion} \label{sec:discussion}

To better understand dependencies within the proposed taxonomy, we perform two levels of analysis: one at the category level (\textit{e.g., GT Value, Generability}, etc.), and another at the subcategory level (\textit{e.g., Binary, Relative, Multivalue, Generalizable, Non-generalizable}, etc.).

\begin{figure}[b]
    \centering
    \includegraphics[width=1.0\textwidth]{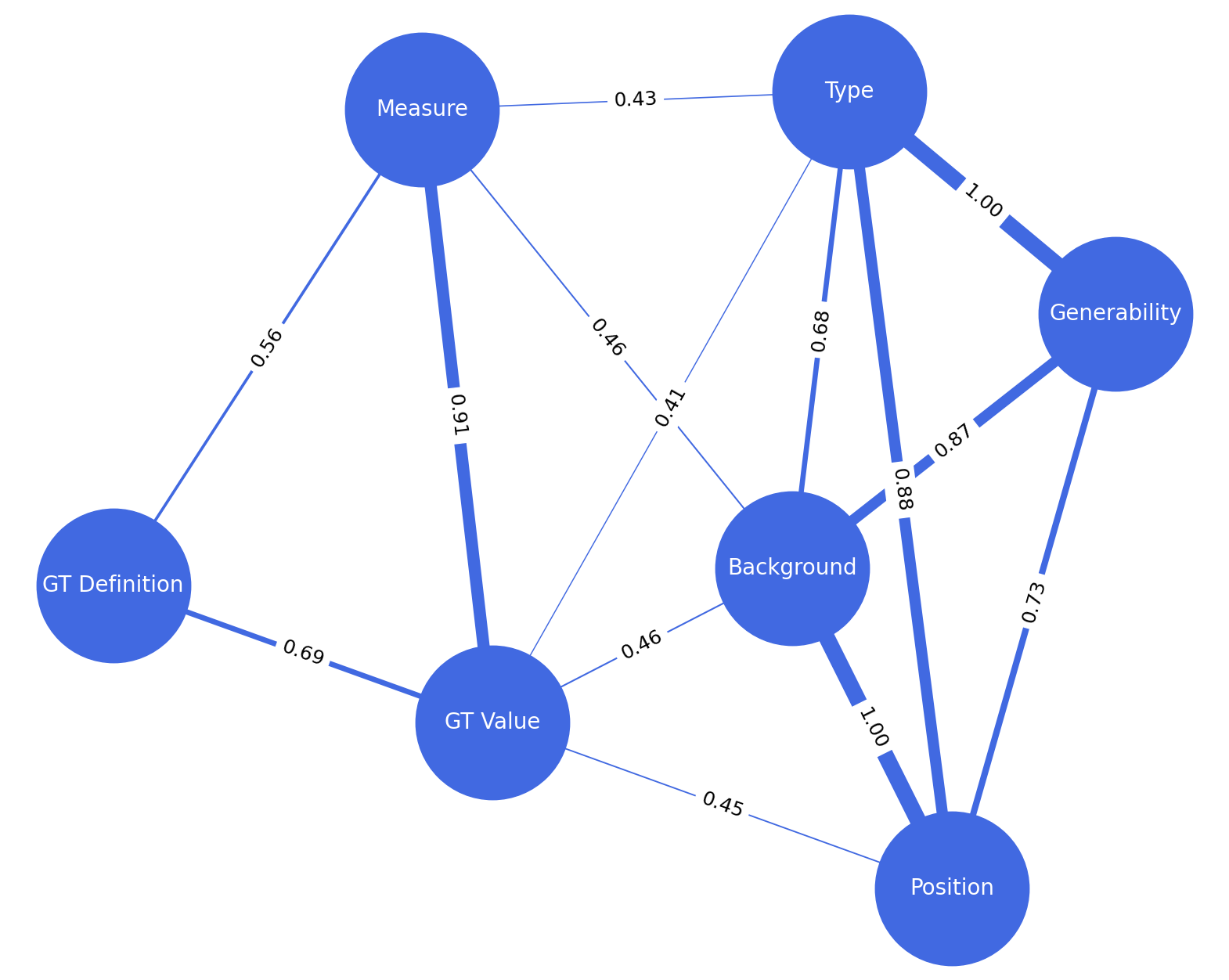}
    \caption{Dependency graph based on Cramér's V. It shows the strength of associations between categories.}
    \label{fig:graph}
\end{figure}

First, we compute Cramér's V, a statistical measure of association between categorical variables that ranges from 0 (no association) to 1 (perfect association). To visualize these associations, we build a dependency graph (see Figure \ref{fig:graph}), where each node represents a category and edges connect those with a Cramér's V above a selected threshold (0.4). The thickness of each edge reflects the strength of the association. This graph provides an overview of how different aspects of evaluation methods, such as the GT definition, background choices, or generability, interact and influence one another.

\begin{figure}[t]
    \centering
    \includegraphics[width=1.0\textwidth]{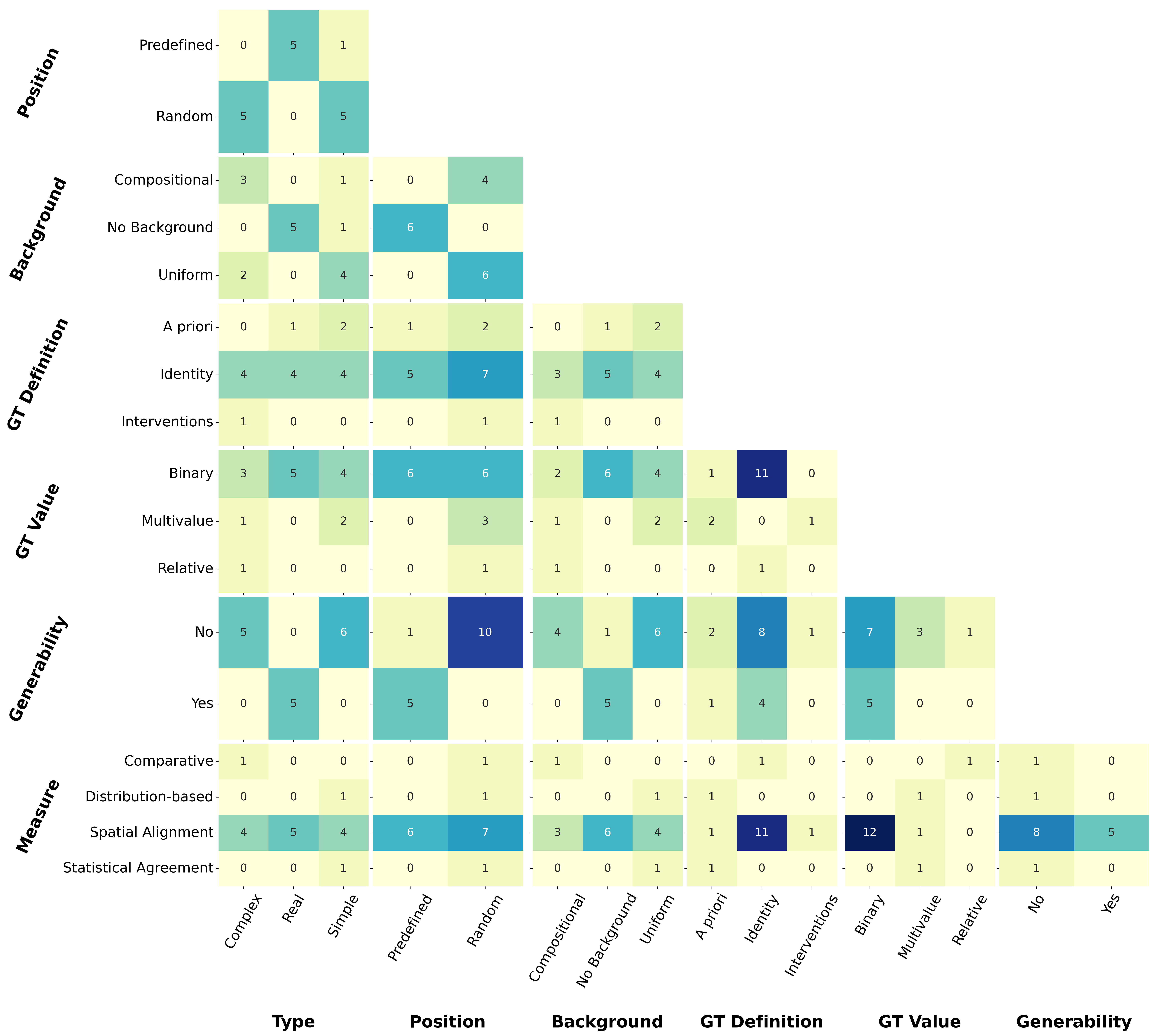}
    \caption{Co-occurrence matrix between subcategories.}
    \label{fig:matrix}
\end{figure}

To further explore the strongest dependencies, we generate a complementary visualization: a pairwise co-occurrence heatmap (see Figure \ref{fig:matrix}). Rows and columns represent subcategories from broader taxonomy categories, and the color intensity of each cell reflects how frequently each pair co-occurs in the survey data (Table \ref{tab:methods-table}). Darker colors indicate more frequent co-occurrence. Let us analyze the fifth strongest dependencies.

\paragraph{Position and Background (Cramér's V = 1.0).}
There is a perfect association between the \textit{position} and the \textit{background} variables. The \textit{no-background} configuration, typically associated with mosaic-based setups, exclusively occurs with \textit{predefined} positioning (six out of six cases, see Figure \ref{fig:matrix}). In contrast, \textit{random} positioning is always paired with either \textit{uniform} or \textit{compositional} backgrounds, and never with \textit{no-background}.

\paragraph{Generability and Type (1.0).} There is also a perfect association between the \textit{generability} of an evaluation method and the \textit{type} of visual elements used in its dataset. For a method to be generalizable, it must not be tied to a specific dataset but should be applicable to real-world datasets and adaptable to new ones. In practice, all generalizable methods in the survey use \textit{real} images as their visual input type. In contrast, all methods using \textit{simple} or \textit{complex} synthetic figures are non-generalizable. These typically rely on pre-designed datasets that limit their portability to other settings. This pattern suggests that real-world image types are a necessary condition for \textit{generalizability}.

\paragraph{GT Value and Measure (0.91).} The choice of evaluation measure is strongly influenced by the type of GT value used. When the GT is a \textit{binary} mask, studies typically apply \textit{Spatial Alignment Metrics} (12 out of 12 in this survey). For \textit{relative GT} values (only one case here), which aim to capture differences in importance across regions, \textit{Comparative} metric is used to evaluate how well the explanation reflects these relative differences. In the case of \textit{multivalue GT} there is no consistent association with a single type of metric.

\paragraph{Type and Position (0.88).}

There is also a dependency between the \textit{type} of visual element and its \textit{position}. All methods that use \textit{random} positioning employ either simple or complex synthetic figures, none of the methods with random positioning use real-world images. In contrast, almost all methods using \textit{real-world} images rely on \textit{predefined} positioning. These are typically mosaic-based methods, where images are placed in fixed layouts to control the GT position. Only one method deviates from this pattern.

\paragraph{Generability and Background (0.87).} There is a relationship between the \textit{generability} category and the \textit{background} category. Specifically, datasets with \textit{no background}, which typically use mosaics, tend to be more generalizable: all five generalizable methods in the survey use \textit{no background}. These methods allow users to generate their own GT, providing greater flexibility. In contrast, datasets with \textit{uniform} or \textit{compositional} backgrounds are typically associated with \textit{non-generalizable} methods, relying on fixed GT datasets and offering less adaptability.

%\paragraph{GT Definition and GT Value (0.69).} The relationship between \textit{GT Definition} and \textit{GT Value} shows a moderate dependency, as the way the GT is defined guides how it is represented (\textit{i.e.,} its value): \textit{binary}, \textit{relative} or \textit{multivalue}. The \textit{identity} definition mostly results in \textit{binary} GT values (11 out of 13). The \textit{a priori} definition is more diverse, yielding both \textit{binary} and \textit{multivalue} values. The \textit{interventions} definition leads to \textit{multivalue} GT values. In summary, identity-based GTs tend to be binary, while more complex definitions like \textit{a priori} and \textit{interventions} require richer representations to capture explanation relevance accurately. 

This analysis shows that methodological choices in SAIG techniques are highly interdependent. Evaluation metrics are strongly constrained by the form of GT values: binary GTs are constently paired with spatial alignment metrics, while more expressive GTs (\textit{e.g.,} multivalue or relative) necessitate alternative evaluation strategies. Design elements such as background type, feature positioning, and visual element type tend to co-occur in consistent patterns, for instance, mosaic-based setups typically combine no-background, predefined positions, and real-world images. Furthemore, generalizable evaluations are associated with real-world images, whereas simple or complex types are strongly linked to non-generalizable setups.  

\input{xai_methods_table}

\subsection{Disagreement in SAIG Evaluations}

It is also important to analyze the results obtained by each SAIG method. Table~\ref{tab:xai_methods} indicates, when specified by the original authors, which XAI methods were considered the best according to their respective evaluation criteria. However, some authors did not specify which XAI method ranked highest in their evaluation.

The data show a clear lack of consensus among the different SAIG methodologies, with various methods being identified as the best across studies. This variation suggests that the performance of XAI methods is highly context-dependent, influenced by factors such as the underlying model and dataset. Furthermore, as already discussed, most methods do not make a comprehensive review of this field prior to their proposal. To the best of our knowledge, this is the first paper to provide such a review of this field. As a result, the lack of consensus has remained largely unrecognised in the analyses of the existing SAIG methods.

Therefore, although one of the main objectives of all the reviewed SAIG methods is to address the limitations of XAI methods and metrics, particularly the disagreement problem~\cite{krishna2024disagreement}, these approaches themselves also exhibit disagreement. Interestingly, an inverse form of consensus can be observed: certain methods are consistently never identified as the best. This recurring underperformance suggests a potential lack of reliability or correctness in some XAI methods.

We expected that any \textit{post-hoc} method would be less faithful to the explanation than an \textit{ante-hoc} approach, and therefore to obtain worse results. We can verify whether this is the case with the two SAIG methods that used \textit{ante-hoc} approaches: Hesse \emph{et al.}~\cite{hesse2023funnybirds} and Fresz \emph{et al.}~\cite{fresz2024classification}. We can see that our expectation, in the study of Fresz \emph{et al.}~\cite{fresz2024classification}, is not true. We consider that this unexpected behavior arises because this SAIG method requires high AI accuracy to serve as a reliable GT.

\section{Conclusions} \label{sec:conclusions}

In this paper, we introduced the concept of SAIG methods as a unified class of machine-centered approaches for evaluating XAI methods. Although such methods have emerged independently in various studies, this work is the first to formally define and categorize them within a common framework. Our proposed taxonomy analyzes the design dimensions of these methods, including GT definition, image composition, evaluation metrics, generability, among others. By doing so, we introduce SAIG as a methodological family within the broader landscape of XAI evaluation. 

Our analysis revealed interdependencies among design decisions in SAIG methods. Furthermore, we identified a lack of consensus in SAIG evaluations. Despite their aim of providing objective evaluations of XAI methods, these approaches often result in divergent conclusions regarding which explanation techniques are the most faithful. This shows that SAIG evaluations, while promising, remain sensitive to design assumptions, task definition, and dataset choices.

XAI methods are widely used not only by developers but also by non-technical users, particularly the feature relevance methods, also called feature attribution methods \cite{calderon-reichart-2025-behalf}. However, the reliability of these methods remains an open question. Before assessing how convincing these explanations are to human users, we argue that it is essential first to evaluate how accurately they reflect the behavior of the underlying model. Since this is precisely the goal of SAIG methods, and given that they are all grounded in the same main idea of generating a GT to evaluate explainability methods, a promising direction for future work is the development of a unified framework that integrates existing SAIG approaches. By leveraging both their shared foundations and methodological differences, such a framework could help address the inconsistencies currently observed across SAIG evaluations and support more robust and comparable assessments of XAI methods.

\section*{Declarations}

\paragraph{Funding}

Anna Arias Duart acknowledges her AI4S fellowship within the “Generación D” initiative by Red.es, Ministerio para la Transformación Digital y de la Función Pública, for talent attraction (C005/24-ED CV1), funded by NextGenerationEU through the Recovery, Transformation and Resilience Plan (PRTR). Gabriel Moyà-Alcover and Miquel Miró-Nicolau contribution is part of the Project PID2023-149079OBI00 funded by MICIU/AEI/10.13039/501100011033 and by ERDF/EU.

\paragraph{Competing Interests}
The authors have no relevant financial or non-financial interests to disclose.

%\cite{ross2017right} %id2 x
%\cite{oramasvisual} %id3 x
%\cite{yang2019benchmarking} %id4
%\cite{shah2021input} %id8
%\cite{guidotti2021evaluating} %id9 x
%\cite{arias2022focus} %id11 x
%\cite{arras2022clevr} %id12 x
%\cite{mamalakis2022investigating} %id13 x
%\cite{rao2022better} %id15 x
%\cite{kim2022sanity} %id18 x
%\cite{hesse2023funnybirds} %id19
%\cite{miro2024assessing} %id20 x
%\cite{bohle2021convolutional} %21 x
%\cite{fresz2024classification} %22 x
%\cite{zhou2022feature} %27
%\cite{tjoa2022quantifying} %29


\bibliography{sn-bibliography.bib}% common bib file

\end{document}

%% file: type.tex
\begin{figure}[htbp]
  \centering
  \begin{tikzpicture}
    \begin{axis}[
        /pgf/number format/1000 sep={},
        width=3.8in,
        height=1.8in,
        scale only axis,
        clip=false,
        separate axis lines,
        xmin=0,
        xmax=4,
        xtick={1,2,3},
        x tick style={draw=none},
        xticklabels={Simple, Complex, Real-World},
        ylabel={Type},
        ymin=0,
        ymax=8,
        grid=major,
        grid style={dashed, gray!20},
        nodes near coords,
        nodes near coords align={vertical},
        axis lines=left,
        every node near coord/.append style={
            font=\small,
            color=black,
            yshift=2pt
        },
        every axis plot/.append style={
          ybar,
          bar width=45pt,
          bar shift=0pt,
          fill
        }
      ]
      \addplot[fill=green1] coordinates {(1,7)};
      \addplot[fill=green2] coordinates {(2,4)};
      \addplot[fill=green3] coordinates {(3,5)};
    \end{axis}
  \end{tikzpicture}
  \caption{Distribution of visual elements by its type.}\label{fig:type}
\end{figure}

%% file: positions.tex
\begin{figure}[htbp]
  \centering
  \begin{tikzpicture}
    \begin{axis}[
        /pgf/number format/1000 sep={},
        width=3.8in,
        height=1.8in,
        scale only axis,
        clip=false,
        separate axis lines,
        xmin=0,
        xmax=3,
        xtick={1,2},
        x tick style={draw=none},
        xticklabels={Random, Predefined},
        ylabel={Position},
        ymin=0,
        ymax=11,
        grid=major,
        grid style={dashed, gray!20},
        nodes near coords,
        nodes near coords align={vertical},
        axis lines=left,
        every node near coord/.append style={
            font=\small,
            color=black,
            yshift=2pt
        },
        every axis plot/.append style={
          ybar,
          bar width=60pt,
          bar shift=0pt,
          fill
        }
      ]
      \addplot[fill=blue1] coordinates {(1,10)};
      \addplot[fill=blue2] coordinates {(2,6)};
    \end{axis}
  \end{tikzpicture}
  \caption{Distribution of visual elements by position. Five out of six articles that use predefined positions rely on mosaics.} \label{fig:positions}
\end{figure}

%% file: background.tex
\begin{figure}[htbp]
  \centering
  \begin{tikzpicture}
    \begin{axis}[
        /pgf/number format/1000 sep={},
        width=3.8in,
        height=1.8in,
        scale only axis,
        clip=false,
        separate axis lines,
        xmin=0,
        xmax=4,
        xtick={1,2,3},
        x tick style={draw=none},
        xticklabels={Uniform, Compositional, No Background},
        ylabel={Background},
        ymin=0,
        ymax=7,
        grid=major,
        grid style={dashed, gray!20},
        nodes near coords,
        nodes near coords align={vertical},
        axis lines=left,
        every node near coord/.append style={
            font=\small,
            color=black,
            yshift=2pt
        },
        every axis plot/.append style={
          ybar,
          bar width=45pt,
          bar shift=0pt,
          fill
        }
      ]
      \addplot[fill=red1] coordinates {(1,6)};
      \addplot[fill=red2] coordinates {(2,5)};
      \addplot[fill=red3] coordinates {(3,5)};
    \end{axis}
  \end{tikzpicture}
  \caption{Types of backgrounds in SAIG datasets.} \label{fig:background}
\end{figure}

%% file: features.tex
\begin{figure}[htbp]
    \centering
    \begin{subfigure}[t]{0.3\textwidth}
        \includegraphics[width=\linewidth]{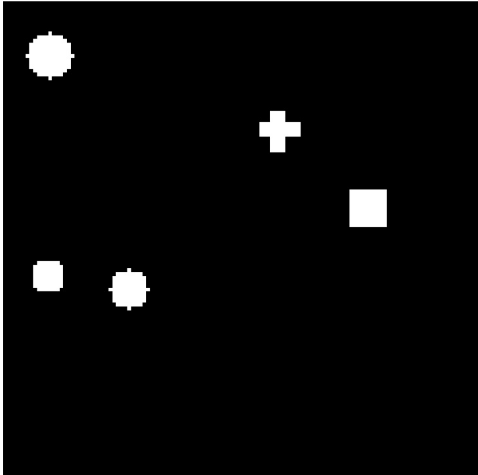}
        \caption{Image from \cite{miro2024assessing}. Simple objects, random position and uniform background.}
    \end{subfigure}
    %\hfill
    \begin{subfigure}[t]{0.3\textwidth}
        \includegraphics[width=\linewidth]{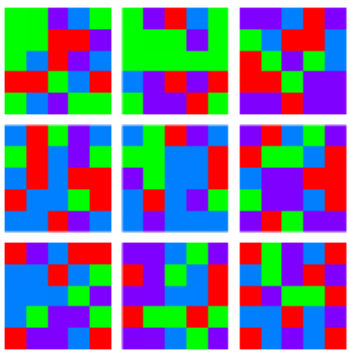}
        \caption{Image from \cite{ross2017right}.  Simple objects, predefined position and no background.}
    \end{subfigure}
    %\hfill
    \begin{subfigure}[t]{0.3\textwidth}
        \includegraphics[width=\linewidth]{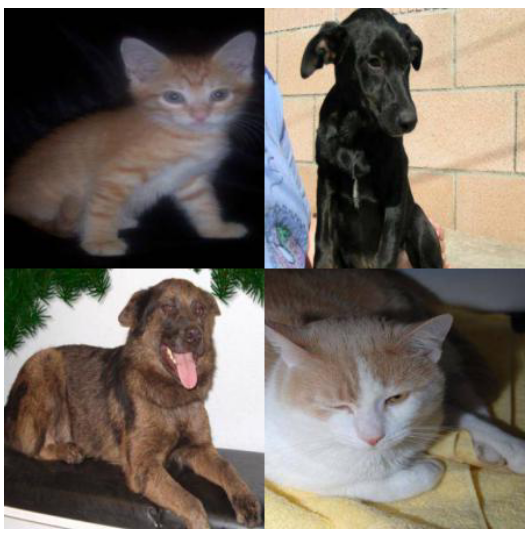}
        \caption{Image from \cite{arias2022focus}. Real world images, predefined position and no background.}
    \end{subfigure}
    %\hfill
    \begin{subfigure}[t]{0.3\textwidth}
        \includegraphics[width=\linewidth]{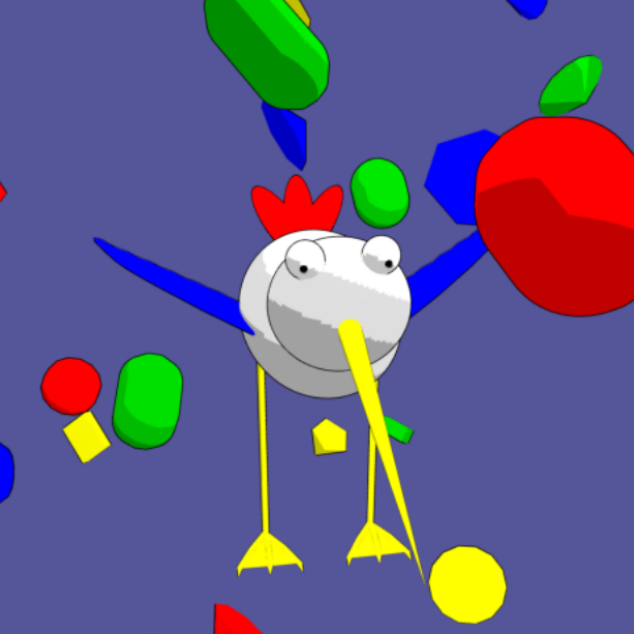}
        \caption{Image from \cite{hesse2023funnybirds}. Complex objects, random position and compositional background.}
    \end{subfigure}
    %\hfill
    \begin{subfigure}[t]{0.3\textwidth}
        \includegraphics[width=\linewidth]{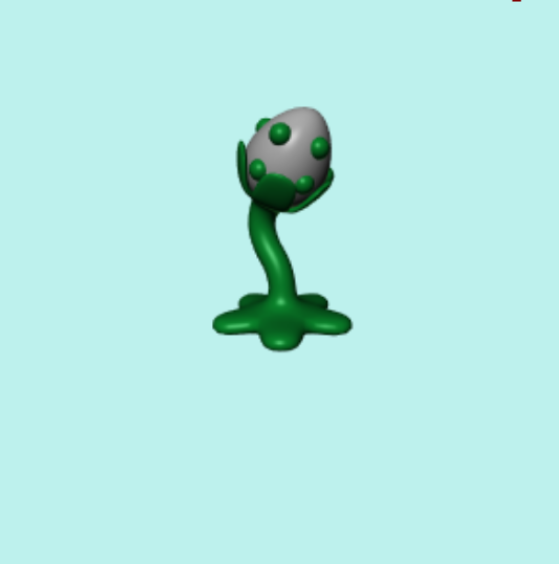}
        \caption{Image from \cite{oramasvisual}. Complex objects, random position, uniform background.}
    \end{subfigure}
    %\hfill
    \begin{subfigure}[t]{0.3\textwidth}
        \includegraphics[width=\linewidth]{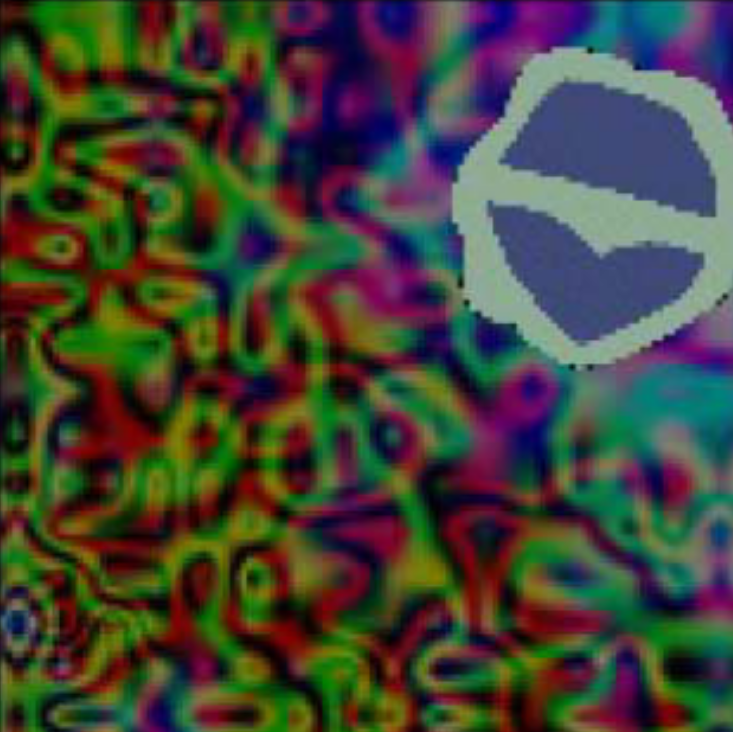}
        \caption{Image from \cite{tjoa2022quantifying}. Simple objects, random position and compositional background.}
    \end{subfigure}

    \caption{Representative examples of the different combinations of object type, position, and background used in the analyzed datasets.}
    \label{fig:galeria}
\end{figure}

%% file: gt_definition.tex
\begin{figure}[htbp]
  \centering
  \begin{tikzpicture}
    \begin{axis}[
        /pgf/number format/1000 sep={},
        width=3.8in,
        height=1.8in,
        scale only axis,
        clip=false,
        separate axis lines,
        xmin=0,
        xmax=4,
        xtick={1,2,3},
        x tick style={draw=none},
        xticklabels={Identity,\textit{A priori}, Interventions},
        ylabel={GT Definition},
        ymin=0,
        ymax=14,
        grid=major,
        grid style={dashed, gray!20},
        nodes near coords,
        nodes near coords align={vertical},
        axis lines=left, % change from 'left' to 'box'
        every node near coord/.append style={
            font=\small,
            color=black,
            yshift=2pt
        },
        every axis plot/.append style={
          ybar,
          bar width=45pt,
          bar shift=0pt,
          fill
        }
      ]
      \addplot[fill=orange1]coordinates {(1,12)};
      \addplot[fill=orange2]coordinates{(2,3)};
      \addplot[fill=orange3]coordinates{(3,1)};
    \end{axis}
  \end{tikzpicture}
  \caption{Distribution of SAIG methods by GT definition categories. The majority use the Identity definition, followed by \textit{A priori}.} \label{fig:gt_definition}
\end{figure}

%% file: gt_value.tex
\begin{figure}[htbp]
  \centering
  \begin{tikzpicture}
    \begin{axis}[
        /pgf/number format/1000 sep={},
        width=3.8in,
        height=1.8in,
        scale only axis,
        clip=false,
        separate axis lines,
        xmin=0,
        xmax=4,
        xtick={1,2,3},
        x tick style={draw=none},
        xticklabels={Binary,Relative,Multivalue},
        ylabel={GT value},
        ymin=0,
        ymax=14,
        grid=major,
        grid style={dashed, gray!20},
        nodes near coords,
        nodes near coords align={vertical},
        axis lines=left, % change from 'left' to 'box'
        every node near coord/.append style={
            font=\small,
            color=black,
            yshift=2pt
        },
        every axis plot/.append style={
          ybar,
          bar width=45pt,
          bar shift=0pt,
          fill
        }
      ]
      \addplot[fill=purple1]coordinates {(1,12)};
      \addplot[fill=purple2]coordinates{(2,1)};
      \addplot[fill=purple3]coordinates{(3,3)};
    \end{axis}
  \end{tikzpicture}
  \caption{Number of SAIG methods by GT value types. The majority of methods use Binary GT values, with fewer employing Multivalue and only one using Relative GT values.}\label{fig:gt_value}
\end{figure}

%% file: portability.tex
\begin{figure}[htbp]
  \centering
  \begin{tikzpicture}
    \begin{axis}[
        /pgf/number format/1000 sep={},
        width=3.2in,
        height=1.8in,
        scale only axis,
        clip=false,
        separate axis lines,
        xmin=0,
        xmax=3,
        xtick={1,2},
        x tick style={draw=none},
        xticklabels={Generalizable, Non-generalizable},
        ylabel={Gerability},
        ymin=0,
        ymax=11,
        grid=major,
        grid style={dashed, gray!20},
        nodes near coords,
        nodes near coords align={vertical},
        axis lines=left,
        every node near coord/.append style={
            font=\small,
            color=black,
            yshift=2pt
        },
        every axis plot/.append style={
          ybar,
          bar width=60pt,
          bar shift=0pt,
          fill
        }
      ]
      \addplot[fill=yellow1] coordinates {(1,5)};
      \addplot[fill=yellow2] coordinates {(2,11)};
    \end{axis}
  \end{tikzpicture}
  \caption{Number of generalizable and non-generalizable methods. The majority (11 out of 16) are non-generalizable.}\label{fig:portability}
\end{figure}

%% file: evaluation.tex
\begin{figure}[htbp]
  \centering
  \begin{tikzpicture}
    \begin{axis}[
        /pgf/number format/1000 sep={},
        width=3.8in,
        height=1.8in,
        scale only axis,
        clip=false,
        separate axis lines,
        xmin=0,
        xmax=5,
        xtick={1,2,3,4},
        x tick style={draw=none},
        xticklabels={
          {\parbox[c][2em][c]{3cm}{\centering Spatial\\Alignment}},
          {\parbox[c][2em][c]{3cm}{\centering Distribution\\based}},
          {\parbox[c][2em][c]{3cm}{\centering Statistical\\Agreement}},
          {\parbox[c][2em][c]{3cm}{\centering Comparative}}
        },
        ylabel={Evaluation measures},
        ymin=0,
        ymax=14,
        grid=major,
        grid style={dashed, gray!20},
        nodes near coords,
        nodes near coords align={vertical},
        axis lines=left,
        every node near coord/.append style={
            font=\small,
            color=black,
            yshift=2pt
        },
        every axis plot/.append style={
          ybar,
          bar width=45pt,
          bar shift=0pt,
          fill
        }
      ]
      \addplot[fill=teal1] coordinates {(1,13)};
      \addplot[fill=teal2] coordinates {(2,1)};
      \addplot[fill=teal3] coordinates {(3,1)};
      \addplot[fill=teal4] coordinates {(4,1)};
    \end{axis}
  \end{tikzpicture}
  \caption{Counts of evaluation measures by category. The majority are Spatial Alignment measures (13), while Distribution-based, Statistical Agreement, and Comparative measures are less frequent, each with a count of 1.}\label{fig:measures}
\end{figure}

%% file: method_table.tex
\begin{landscape}
\begin{table}[h!]
\centering
\caption{Overview of SAIG approaches and their evaluation properties. The colored headers are linked to the sections where each dimension is discussed in detail.}
\renewcommand{\arraystretch}{2}  % aumenta el espacio vertical entre filas
\begin{tabular}{
  >{\centering\arraybackslash}m{0.7cm}   % icon
  >{\centering\arraybackslash}m{2.3cm} % type
  >{\centering\arraybackslash}m{2.3cm} % position
  >{\centering\arraybackslash}m{2.3cm} % background
  >{\centering\arraybackslash}m{2.3cm} % GT Definition
  >{\centering\arraybackslash}m{2.3cm} % GT Value
  >{\centering\arraybackslash}m{2.3cm} % Generability
  >{\centering\arraybackslash}m{3cm}   % Measure
}
\toprule
\parbox[c][0.7cm][t]{0.5cm}{\vspace{0.1cm}\includegraphics[height=0.5cm]{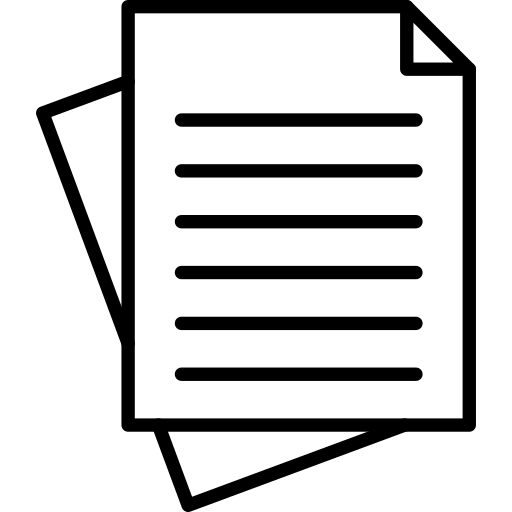}} & 
\multicolumn{3}{c}{\textbf{Image Features}} & 
\textbf{\hyperlink{gt-definition}{\colorbox{orange3}{GT Definition}}} & 
\textbf{\hyperlink{gt-value}{\colorbox{purple3}{GT Value}}} & 
\textbf{\hyperlink{generability-section}{\colorbox{yellow2}{Generability}}} & 
\textbf{\hyperlink{evaluation-section}{\colorbox{teal1}{Measure}}} \\
\cmidrule(lr){2-4}
& \textbf{\hyperlink{type-section}{\colorbox{green3}{Type}}
} & \textbf{\hyperlink{position-section}{\colorbox{blue1}{Position}}} & \textbf{\hyperlink{background-section}{\colorbox{red3}{Background}}} & & & & \\
\midrule
\cite{ross2017right} & Simple& Predefined & No Background & Identity & Binary & No & Spatial Alignment  \\
\cite{oramasvisual} & Complex & Random & Uniform & Identity  & Binary & No & Spatial Alignment \\
\cite{yang2019benchmarking} & Complex & Random & Compositional & Identity  & Relative & No & Comparative \\
\cite{shah2021input} & Real& Predefined & No Background & \textit{A priori} & Binary & Yes & Spatial Alignment \\
\cite{guidotti2021evaluating} & Simple&  Random & Uniform & Identity  & Binary & No & Spatial Alignment \\
\cite{arias2022focus} & Real & Predefined & No Background & Identity & Binary & Yes & Spatial Alignment \\
\cite{arras2022clevr} & Complex& Random & Uniform & Identity  &  Binary & No & Spatial Alignment \\
\cite{mamalakis2022investigating} & Simple & Random & Uniform & \textit{A priori} & Multivalue (sign) & No & Statistical Agreement \\
\cite{rao2022better} & Real & Predefined & No Background & Identity & Binary & Yes & Spatial Alignment \\
\cite{kim2022sanity} & Simple & Random & Uniform & Identity & Binary & No & Spatial Alignment \\
\cite{hesse2023funnybirds} & Complex & Random & Compositional & Interventions & Multivalue & No & Spatial Alignment \\
\cite{miro2024assessing} & Simple & Random & Uniform & \textit{A priori} & Multivalue & No & Distribution-based \\
\cite{bohle2021convolutional} & Real & Predefined & No Background & Identity & Binary & Yes & Spatial Alignment \\
\cite{fresz2024classification} & Real & Predefined & No Background & Identity & Binary & Yes & Spatial Alignment \\
\cite{zhou2022feature} & Complex & Random & Compositional & Identity &  Binary & No & Spatial Alignment \\ 
\cite{tjoa2022quantifying} & Simple & Random & Compositional & Identity & Binary & No & Spatial Alignment \\
\bottomrule
\end{tabular}
\label{tab:methods-table}
\end{table}
\end{landscape}

%% file: xai_methods_table.tex
\begin{landscape}

\begin{table}[p]
    \centering
    \caption{XAI methods used in each SAIG method. The first column grouped the methods according to the proposal from Speith~\cite{speith2022review}, the second according to their output type, and the latter according to their inner working. In green, the best XAI method according to each SAIG article.}
    \addtolength{\tabcolsep}{-0.35em}
    \begin{tabular}{llllccccccccccccccccc}
    \toprule
        Speith~\cite{speith2022review} & Output & \makecell[l]{Inner \\ working} &  & \cite{ross2017right} & \cite{oramasvisual} & \cite{yang2019benchmarking} & \cite{shah2021input} & \cite{guidotti2021evaluating} & \cite{arias2022focus} & \cite{arras2022clevr} & \cite{mamalakis2022investigating} & \cite{rao2022better} & \cite{kim2022sanity} & \cite{hesse2023funnybirds} & \cite{miro2024assessing} & \cite{bohle2021convolutional} & \cite{fresz2024classification} & \cite{zhou2022feature} & \cite{tjoa2022quantifying} & Total \\ \midrule
        \multirow{4}{*}{\textit{Ante-hoc}} & \makecell[l]{Expl. by concepts}  & &  BagNet~\cite{brendel2018approximating} & ~ & ~ & ~ & ~ & ~ & ~ & ~ & ~ & ~ & ~ & \colorbox{green3}{\checkmark} & ~ & ~ & ~ & ~ & ~ & 1 \\ \arrayrulecolor{black!30}\cmidrule(lr){2-21}
        & \makecell[l]{Expl. by examples}  & &   ProtoPNet~\cite{chen2019looks} & ~ & ~ & ~ & ~ & ~ & ~ & ~ & ~ & ~ & ~ & \checkmark & ~ & ~ & ~ & ~ & ~ & 1 \\ \cmidrule(lr){2-21}
        &  \multirow{2}{*}{\makecell[l]{Feature Relevance}} & &   B-cos~\cite{bohle2024bcos} & ~ & ~ & ~ & ~ & ~ & ~ & ~ & ~ & ~ & ~ & \checkmark & ~ & ~ & \checkmark & ~ & ~ & 2 \\ 
        &   &  &   X-DNN~\cite{hesse2021fast} & ~ & ~ & ~ & ~ & ~ & ~ & ~ & ~ & ~ & ~ & \checkmark & ~ & ~ & ~ & ~ & ~ & 1 \\   \midrule

        \multirow{28}{*}{\textit{Post-hoc}} & \multirow{27}{*}{\makecell[l]{Feature \\ Relevance}} & \multirow{14}{*}{Backprop.} &    Deconvnet~\cite{zeiler2014visualizing} & ~ & \checkmark & ~ & ~ & ~ & ~ & \checkmark & ~ & ~ & \checkmark & ~ & \checkmark & ~ & ~ & ~ & \checkmark & 5 \\ 
        &&&   DeepLIFT~\cite{shrikumar2019learning} & ~ & ~ & ~ & ~ & ~ & ~ & ~ & ~ & ~ & \colorbox{green3}{\checkmark} & ~ & \checkmark & \checkmark & ~ & ~ & \checkmark & 4 \\ 
        &&&   DeepTaylor~\cite{montavon2018methods} & ~ & ~ & ~ & ~ & ~ & ~ & ~ & \checkmark & ~ & \checkmark & ~ & ~ & ~ & ~ & ~ & ~ & 2 \\ 
        &&&   Excit. Backprop.~\cite{zhang2018topdown}  & ~ & ~ & ~ & ~ & ~ & ~ & \checkmark & ~ & ~ & ~ & ~ & ~ & ~ & ~ & ~ & ~ & 1 \\ 
        &&&   Gradient~\cite{simonyan2014deep} & \checkmark & ~ & \colorbox{green3}{\checkmark} & ~ & ~ & ~ & \checkmark & \checkmark & \checkmark & \checkmark & ~ & \checkmark & \checkmark & ~ & \checkmark & \checkmark & 10 \\ 
        &&&   GBP~\cite{springenberg2014striving} & ~ & \checkmark & \checkmark & \checkmark & ~ & ~ & \checkmark & ~ & \checkmark & \checkmark & ~ & \checkmark & ~ & ~ & ~ & \checkmark & 8 \\ 
        &&&   Input X Grad.~\cite{shrikumar2019learning} & ~ & ~ & \checkmark & \checkmark & ~ & ~ & \checkmark & \checkmark & \checkmark & \checkmark & \checkmark & ~ & \checkmark & ~ & ~ & \checkmark & 9 \\ 
        &&&   Int. Gradients~\cite{sundararajan2017axiomatic} & ~ & ~ & \checkmark & \checkmark & ~ & \checkmark & \checkmark & \checkmark & \checkmark & \checkmark & \checkmark & \checkmark & \checkmark & \checkmark & ~ & ~ & 11 \\ 
        &&&   LRP~\cite{bach2015pixelwise} & ~ & ~ & ~ & \checkmark & ~ & \checkmark & \colorbox{green3}{\checkmark} & \checkmark & ~ & \checkmark & ~ & \colorbox{green3}{\checkmark} & ~ & \colorbox{green3}{\checkmark} & ~ & ~ & 7 \\ 
        &&&   Chefer LRP~\cite{chefer2021transformer} & ~ & ~ & ~ & ~ & ~ & ~ & ~ & ~ & ~ & ~ & \checkmark & ~ & ~ & ~ & ~ & ~ & 1 \\
        &&&   PatternAttr.~\cite{kindermans2018learning} & ~ & ~ & ~ & ~ & ~ & ~ & ~ & \checkmark & ~ & ~ & ~ & ~ & ~ & ~ & ~ & ~ & 1 \\ 
        &&&   PatternNet~\cite{kindermans2018learning} & ~ & ~ & ~ & ~ & ~ & ~ & ~ & \checkmark & ~ & ~ & ~ & ~ & ~ & ~ & ~ & ~ & 1 \\ 
        &&&   SmoothGrad~\cite{smilkov2017smoothgrad} & ~ & ~ & \checkmark & \checkmark & ~ & \checkmark & \checkmark & \checkmark & ~ & \checkmark & ~ & \checkmark & ~ & \checkmark & \checkmark & ~ & 9 \\
        &&&   VarGrad~\cite{adebayo2018local} & ~ & ~ & ~ & ~ & ~ & ~ & \checkmark & ~ & ~ & ~ & ~ & ~ & ~ & ~ & ~ & ~ & 1 \\ \cmidrule(lr){3-21}

        &&\multirow{7}{*}{CAM} &   AblationCAM~\cite{desai2020ablationcam} & ~ & ~ & ~ & ~ & ~ & ~ & ~ & ~ & \checkmark & ~ & ~ & \checkmark & ~ & ~ & ~ & ~ & 2 \\ 
        &&&   GradCAM~\cite{selvaraju2020gradcam} & ~ & \checkmark & \colorbox{green3}{\checkmark} & ~ & ~ & \colorbox{green3}{\checkmark} & \checkmark & ~ & \checkmark & \checkmark & \checkmark & \checkmark & \checkmark & \checkmark & \checkmark & ~ & 11 \\ 
        &&&   Grad-CAM++~\cite{chattopadhay2018gradcam} & ~ & \checkmark & ~ & ~ & ~ & \checkmark & ~ & ~ & \checkmark & ~ & ~ & \checkmark & ~ & \checkmark & ~ & ~ & 5 \\ 
        &&&   Guided GradCAM~\cite{selvaraju2020gradcam} & ~ & \checkmark & ~ & ~ & ~ & ~ & \checkmark & ~ & ~ & ~ & ~ & ~ & ~ & ~ & ~ & \checkmark & 3 \\ 
        &&&   Layer-CAM~\cite{jiang2021layercam} & ~ & ~ & ~ & ~ & ~ & ~ & ~ & ~ & \checkmark & ~ & ~ & ~ & ~ & ~ & ~ & ~ & 1 \\ 
        &&&   Score-CAM~\cite{wang2020scorecam} & ~ & ~ & ~ & ~ & ~ & ~ & ~ & ~ & \checkmark & ~ & ~ & \checkmark & ~ & ~ & ~ & ~ & 2 \\ 
        &&&   SIDU~\cite{muddamsetty2022visual} & ~ & ~ & ~ & ~ & ~ & ~ & ~ & ~ & ~ & ~ & ~ & \checkmark & ~ & ~ & ~ & ~ & 1 \\  \cmidrule(lr){3-21} % thinner than default \midrule

        &&\multirow{6}{*}{Perturb.} &   LIME~\cite{ribeiro2016why} & \checkmark & ~ & ~ & ~ & \colorbox{green3}{\checkmark} & \checkmark & ~ & ~ & ~ & ~ & \checkmark & \checkmark & \checkmark & \checkmark & \checkmark & ~ & 8 \\ 

        &&&   MAPLE~\cite{plumb2018model} & ~ & ~ & ~ & ~ & \checkmark & ~ & ~ & ~ & ~ & ~ & ~ & ~ & ~ & ~ & ~ & ~ & 1 \\ 
        &&&   Occlusion sens.~\cite{zeiler2014visualizing} & ~ & ~ & ~ & \checkmark & ~ & ~ & ~ & ~ & \checkmark & ~ & ~ & \checkmark & \checkmark & ~ & ~ & ~ & 4 \\ 
        &&&   RISE~\cite{petsiuk2018rise} & ~ & ~ & ~ & ~ & ~ & ~ & ~ & ~ & \checkmark & ~ & \checkmark & \checkmark & \checkmark & ~ & ~ & ~ & 4 \\ 
        &&&   Rollout~\cite{abnar2020quantifying}  & ~ & ~ & ~ & ~ & ~ & ~ & ~ & ~ & ~ & ~ & \checkmark & ~ & ~ & ~ & ~ & ~ & 1 \\ 
        &&&   SHAP~\cite{lundberg2017unified} & ~ & ~ & ~ & ~ & \colorbox{green3}{\checkmark} & ~ & ~ & \checkmark & ~ & \checkmark & ~ & \checkmark & ~ & \colorbox{green3}{\checkmark} & \colorbox{green3}{\checkmark} & \checkmark & 7 \\ \cmidrule(lr){2-21}

        & Expl. by concept &  &   TCAV~\cite{kim2018interpretability} & ~ & ~ & \colorbox{green3}{\checkmark} & ~ & ~ & ~ & ~ & ~ & ~ & ~ & ~ & ~ & ~ & ~ & ~ & ~ & 1 \\ 
        \arrayrulecolor{black}\bottomrule
    \end{tabular}
    \label{tab:xai_methods}
\end{table}

\end{landscape}

%% file: sn-bibliography.bib
@inproceedings{oramasvisual,
  title={Visual Explanation by Interpretation: Improving Visual Feedback Capabilities of Deep Neural Networks},
  author={Oramas, Jose and Wang, Kaili and Tuytelaars, Tinne},
  booktitle={International Conference on Learning Representations},
  year={2018}
}

@inproceedings{ross2017right,
  title={Right for the Right Reasons: Training Differentiable Models by Constraining their Explanations},
  author={Ross, Andrew Slavin and Hughes, Michael C and Doshi-Velez, Finale},
  booktitle={Proceedings of the Twenty-Sixth International Joint Conference on Artificial Intelligence},
  pages={2662--2670},
  year={2017},
}

@article{yang2019benchmarking,
  title={Benchmarking attribution methods with relative feature importance},
  author={Yang, Mengjiao and Kim, Been},
  journal={arXiv preprint arXiv:1907.09701},
  year={2019}
}

@article{shah2021input,
  title={Do input gradients highlight discriminative features?},
  author={Shah, Harshay and Jain, Prateek and Netrapalli, Praneeth},
  journal={Advances in Neural Information Processing Systems},
  volume={34},
  pages={2046--2059},
  year={2021}
}

@article{guidotti2021evaluating,
  title={Evaluating local explanation methods on ground truth},
  author={Guidotti, Riccardo},
  journal={Artificial Intelligence},
  volume={291},
  pages={103428},
  year={2021},
  publisher={Elsevier}
}

@inproceedings{arias2022focus,
  title={Focus! rating xai methods and finding biases},
  author={Arias-Duart, Anna and Par{\'e}s, Ferran and Garcia-Gasulla, Dario and Gimenez-Abalos, Victor},
  booktitle={2022 IEEE International Conference on Fuzzy Systems (FUZZ-IEEE)},
  pages={1--8},
  year={2022},
  organization={IEEE}
}

@article{arras2022clevr,
  title={CLEVR-XAI: A benchmark dataset for the ground truth evaluation of neural network explanations},
  author={Arras, Leila and Osman, Ahmed and Samek, Wojciech},
  journal={Information Fusion},
  volume={81},
  pages={14--40},
  year={2022},
  publisher={Elsevier}
}

@article{mamalakis2022investigating,
  title={Investigating the fidelity of explainable artificial intelligence methods for applications of convolutional neural networks in geoscience},
  author={Mamalakis, Antonios and Barnes, Elizabeth A and Ebert-Uphoff, Imme},
  journal={Artificial Intelligence for the Earth Systems},
  volume={1},
  number={4},
  pages={e220012},
  year={2022}
}

@inproceedings{kim2022sanity,
  title={Sanity Simulations for Saliency Methods},
  author={Kim, Joon Sik and Plumb, Gregory and Talwalkar, Ameet},
  booktitle={International Conference on Machine Learning},
  pages={11173--11200},
  year={2022},
  organization={PMLR}
}

@article{miro2024assessing,
  title={Assessing fidelity in xai post-hoc techniques: A comparative study with ground truth explanations datasets},
  author={Mir{\'o}-Nicolau, Miquel and Jaume-i-Cap{\'o}, Antoni and Moy{\`a}-Alcover, Gabriel},
  journal={Artificial Intelligence},
  volume={335},
  pages={104179},
  year={2024},
  publisher={Elsevier}
}

@inproceedings{hesse2023funnybirds,
  title={Funnybirds: A synthetic vision dataset for a part-based analysis of explainable ai methods},
  author={Hesse, Robin and Schaub-Meyer, Simone and Roth, Stefan},
  booktitle={Proceedings of the IEEE/CVF International Conference on Computer Vision},
  pages={3981--3991},
  year={2023}
}

@inproceedings{bohle2021convolutional,
  title={Convolutional dynamic alignment networks for interpretable classifications},
  author={Bohle, Moritz and Fritz, Mario and Schiele, Bernt},
  booktitle={Proceedings of the IEEE/CVF Conference on Computer Vision and Pattern Recognition},
  pages={10029--10038},
  year={2021}
}

@inproceedings{fresz2024classification,
  title={Classification metrics for image explanations: towards building reliable XAI-evaluations},
  author={Fresz, Benjamin and L{\"o}rcher, Lena and Huber, Marco},
  booktitle={Proceedings of the 2024 ACM Conference on Fairness, Accountability, and Transparency},
  pages={1--19},
  year={2024}
}

@inproceedings{zhou2022feature,
  title={Do feature attribution methods correctly attribute features?},
  author={Zhou, Yilun and Booth, Serena and Ribeiro, Marco Tulio and Shah, Julie},
  booktitle={Proceedings of the AAAI conference on artificial intelligence},
  volume={36},
  number={9},
  pages={9623--9633},
  year={2022}
}

@article{tjoa2022quantifying,
  title={Quantifying explainability of saliency methods in deep neural networks with a synthetic dataset},
  author={Tjoa, Erico and Guan, Cuntai},
  journal={IEEE Transactions on artificial Intelligence},
  volume={4},
  number={4},
  pages={858--870},
  year={2022},
  publisher={IEEE}
}

@article{krizhevsky2012imagenet,
  title={Imagenet classification with deep convolutional neural networks},
  author={Krizhevsky, Alex and Sutskever, Ilya and Hinton, Geoffrey E},
  journal={Advances in neural information processing systems},
  volume={25},
  year={2012}
}

@inproceedings{biecek2024position,
  title = {Position: {Explain} to {Question} Not to {Justify} },
  booktitle = {Proceedings of the 41st {International Conference} on {Machine Learning} },
  author = {Biecek, Przemyslaw and Samek, Wojciech},
  date = {2024-07-08},
  pages = {3996--4006},
  eventtitle = {International {Conference} on {Machine Learning}}
}

@article{nauta2023anecdotal,
  title = {From {Anecdotal Evidence} to {Quantitative Evaluation Methods}: {A Systematic Review} on {Evaluating Explainable AI} },
  author = {Nauta, Meike and Trienes, Jan and Pathak, Shreyasi and Nguyen, Elisa and Peters, Michelle and Schmitt, Yasmin and Schlötterer, Jörg and Van Keulen, Maurice and Seifert, Christin},
  date = {2023-12-31},
  journaltitle = {ACM Computing Surveys},
  volume = {55},
  pages = {1--42},
  issue = {13s}
}

@misc{amengual-alcover2025evaluation,
  title = {Towards an {Evaluation Framework} for {Explainable Artificial Intelligence Systems} for {Health} and {Well-being} },
  author = {Amengual-Alcover, Esperança and Jaume-i-Capó, Antoni and Miró-Nicolau, Miquel and Moyà-Alcover, Gabriel and Paniza-Fullana, Antonia},
  date = {2025-04-11},
  eprint = {2504.08552},
  eprinttype = {arXiv},
  eprintclass = {cs},
  pubstate = {prepublished}
}

@article{bodria2023benchmarking,
  title = {Benchmarking and Survey of Explanation Methods for Black Box Models},
  author = {Bodria, Francesco and Giannotti, Fosca and Guidotti, Riccardo and Naretto, Francesca and Pedreschi, Dino and Rinzivillo, Salvatore},
  date = {2023-09},
  journaltitle = {Data Mining and Knowledge Discovery},
  volume = {37},
  number = {5},
  pages = {1719--1778},
}

@incollection{doshi-velez2018considerations,
  title = {Considerations for {Evaluation} and {Generalization} in {Interpretable Machine Learning} },
  booktitle = {Explainable and {Interpretable Models} in {Computer Vision} and {Machine Learning} },
  author = {Doshi-Velez, Finale and Kim, Been},
  date = {2018},
  pages = {3--17},
  location = {Cham},
  isbn = {978-3-319-98130-7 978-3-319-98131-4}
}

@article{vilone2021notionsa,
  title = {Notions of Explainability and Evaluation Approaches for Explainable Artificial Intelligence},
  author = {Vilone, Giulia and Longo, Luca},
  date = {2021-12},
  journaltitle = {Information Fusion},
  volume = {76},
  pages = {89--106},
}

@article{bach2015pixelwise,
  title = {On {Pixel-Wise Explanations} for {Non-Linear Classifier Decisions} by {Layer-Wise Relevance Propagation} },
  author = {Bach, Sebastian and Binder, Alexander and Montavon, Grégoire and Klauschen, Frederick and Müller, Klaus-Robert and Samek, Wojciech},
  date = {2015-07-10},
  journaltitle = {PLOS ONE},
  volume = {10},
  number = {7},
  pages = {e0130140},
}

@article{samek2017evaluating,
  title = {Evaluating the {Visualization} of {What} a {Deep Neural Network Has Learned} },
  author = {Samek, Wojciech and Binder, Alexander and Montavon, Gregoire and Lapuschkin, Sebastian and Muller, Klaus-Robert},
  date = {2017-11},
  journaltitle = {IEEE Transactions on Neural Networks and Learning Systems},
  volume = {28},
  number = {11},
  eprint = {27576267},
  eprinttype = {pubmed},
  pages = {2660--2673},
}

@inproceedings{yeh2019infidelity,
  title = {On the (in)Fidelity and Sensitivity of Explanations},
  booktitle = {Advances in Neural Information Processing Systems},
  author = {Yeh, Chih-Kuan and Hsieh, Cheng-Yu and Suggala, Arun and Inouye, David I and Ravikumar, Pradeep K},
  date = {2019},
  volume = {32}
}

@inproceedings{agarwal2022rethinking,
  title = {Rethinking {Stability} for {Attribution-based Explanations} },
  author = {Agarwal, Chirag and Johnson, Nari and Pawelczyk, Martin and Krishna, Satyapriya and Saxena, Eshika and Zitnik, Marinka and Lakkaraju, Himabindu},
  date = {2022-03-25},
  eventtitle = {{ICLR} 2022 {Workshop} on {PAIR}\{\textbackslash textasciicircum\}{2Struct}: {Privacy}, {Accountability}, {Interpretability}, {Robustness}, {Reasoning} on {Structured Data}}
}

@misc{alvarez-melis2018robustness,
  title = {On the {Robustness} of {Interpretability Methods} },
  author = {Alvarez-Melis, David and Jaakkola, Tommi S.},
  date = {2018-06-20},
  eprint = {1806.08049},
  eprinttype = {arXiv},
  eprintclass = {cs, stat},
  pubstate = {prepublished}
}

@article{montavon2018methods,
  title = {Methods for Interpreting and Understanding Deep Neural Networks},
  author = {Montavon, Grégoire and Samek, Wojciech and Müller, Klaus-Robert},
  date = {2018-02},
  journaltitle = {Digital Signal Processing},
  volume = {73},
  pages = {1--15},
}

@article{adadi2018peeking,
  title = {Peeking {Inside} the {Black-Box}: {A Survey} on {Explainable Artificial Intelligence} ({XAI})},
  author = {Adadi, Amina and Berrada, Mohammed},
  date = {2018},
  journaltitle = {IEEE Access},
  volume = {6},
  pages = {52138--52160},
}

@article{barredoarrieta2020explainable,
  title = {Explainable {Artificial Intelligence} ({XAI}): {Concepts}, Taxonomies, Opportunities and Challenges toward Responsible {AI} },
  author = {Barredo Arrieta, Alejandro and Díaz-Rodríguez, Natalia and Del Ser, Javier and Bennetot, Adrien and Tabik, Siham and Barbado, Alberto and Garcia, Salvador and Gil-Lopez, Sergio and Molina, Daniel and Benjamins, Richard and Chatila, Raja and Herrera, Francisco},
  date = {2020-06},
  journaltitle = {Information Fusion},
  volume = {58},
  pages = {82--115},
}

@article{longo2024explainable,
  title = {Explainable {Artificial Intelligence} ({XAI}) 2.0: {A} Manifesto of Open Challenges and Interdisciplinary Research Directions},
  author = {Longo, Luca and Brcic, Mario and Cabitza, Federico and Choi, Jaesik and Confalonieri, Roberto and Ser, Javier Del and Guidotti, Riccardo and Hayashi, Yoichi and Herrera, Francisco and Holzinger, Andreas and Jiang, Richard and Khosravi, Hassan and Lecue, Freddy and Malgieri, Gianclaudio and Páez, Andrés and Samek, Wojciech and Schneider, Johannes and Speith, Timo and Stumpf, Simone},
  date = {2024-06},
  journaltitle = {Information Fusion},
  volume = {106},
  pages = {102301},
}

@inproceedings{speith2022review,
  title = {A {Review} of {Taxonomies} of {Explainable Artificial Intelligence} ({XAI}) methods},
  booktitle = {2022 {ACM Conference} on {Fairness}, {Accountability}, and {Transparency} },
  author = {Speith, Timo},
  date = {2022-06-21},
  pages = {2239--2250},
  location = {Seoul Republic of Korea},
  eventtitle = {{FAccT} '22: 2022 {ACM Conference} on {Fairness}, {Accountability}, and {Transparency} },
  isbn = {978-1-4503-9352-2}
}

@article{chattopadhay2018gradcam,
  title = {Grad-{CAM}++: {Generalized} Gradient-Based Visual Explanations for Deep Convolutional Networks},
  author = {Chattopadhay, Aditya and Sarkar, Anirban and Howlader, Prantik and Balasubramanian, Vineeth N.},
  date = {2018},
  journaltitle = {Proceedings - 2018 IEEE Winter Conference on Applications of Computer Vision, WACV 2018},
  volume = {2018-Janua},
  pages = {839--847},
  isbn = {9781538648865}
}

@article{ribeiro2016why,
  title = {"{Why} Should i Trust You?" {Explaining} the Predictions of Any Classifier},
  author = {Ribeiro, Marco Tulio and Singh, Sameer and Guestrin, Carlos},
  date = {2016},
  journaltitle = {Proceedings of the ACM SIGKDD International Conference on Knowledge Discovery and Data Mining},
  volume = {13-17-Augu},
  eprint = {1602.04938},
  eprinttype = {arXiv},
  pages = {1135--1144},
  isbn = {9781450342322}
}

@misc{simonyan2014deep,
  title = {Deep {Inside Convolutional Networks}: {Visualising Image Classification Models} and {Saliency Maps} },
  author = {Simonyan, Karen and Vedaldi, Andrea and Zisserman, Andrew},
  date = {2014-04-19},
  eprint = {1312.6034},
  eprinttype = {arXiv},
  eprintclass = {cs},
  pubstate = {prepublished}
}

@report{zeiler2014visualizing,
  title = {Visualizing and {Understanding Convolutional Networks} },
  author = {Zeiler, Matthew D and Fergus, Rob},
  date = {2014}
}

@article{zhou2016learning,
  title = {Learning {Deep Features} for {Discriminative Localization} },
  author = {Zhou, Bolei and Khosla, Aditya and Lapedriza, Agata and Oliva, Aude and Torralba, Antonio},
  date = {2016-08},
  journaltitle = {Proceedings of the IEEE Conference on Computer Vision and Pattern Recognition (CVPR)},
  volume = {2004},
  number = {1},
  pages = {2921--2929}
}

@article{krishna2024disagreement,
  title = {The {Disagreement Problem} in {Explainable Machine Learning}: {A Practitioner}’s {Perspective} },
  author = {Krishna, Satyapriya and Han, Tessa and Gu, Alex and Wu, Steven and Jabbari, Shahin and Lakkaraju, Himabindu},
  date = {2024-02-06},
  journaltitle = {Transactions on Machine Learning Research}
}

@article{hedstrom2023metaevaluation,
  title = {The {Meta-Evaluation Problem} in {Explainable AI}: {Identifying Reliable Estimators} with {MetaQuantus} },
  author = {Hedström, Anna and Bommer, Philine Lou and Wickstrøm, Kristoffer Knutsen and Samek, Wojciech and Lapuschkin, Sebastian and Höhne, Marina MC},
  date = {2023-02-17},
  journaltitle = {Transactions on Machine Learning Research}
}

@inproceedings{miro-nicolau2024metaevaluating,
  title = {Meta-Evaluating {Stability Measures}: {MAX-Sensitivity} and~{AVG-Sensitivity} },
  booktitle = {Explainable {Artificial Intelligence} },
  author = {Miró-Nicolau, Miquel and Jaume-i-Capó, Antoni and Moyà-Alcover, Gabriel},
  date = {2024},
  pages = {356--369},
  location = {Cham},
  isbn = {978-3-031-63787-2}
}

@article{miro-nicolau2025comprehensive,
  title = {A Comprehensive Study on Fidelity Metrics for {XAI} },
  author = {Miró-Nicolau, Miquel and Jaume-i-Capó, Antoni and Moyà-Alcover, Gabriel},
  date = {2025-01-01},
  journaltitle = {Information Processing \& Management},
  volume = {62},
  number = {1},
  pages = {103900},
}

@article{tomsett2020sanity,
  title = {Sanity {Checks} for {Saliency Metrics} },
  author = {Tomsett, Richard and Harborne, Dan and Chakraborty, Supriyo and Gurram, Prudhvi and Preece, Alun},
  date = {2020-04-03},
  journaltitle = {Proceedings of the AAAI Conference on Artificial Intelligence},
  volume = {34},
  number = {04},
  eprint = {1912.01451},
  eprinttype = {arXiv},
  pages = {6021--6029},
  isbn = {9781577358350}
}

@article{agarwal2022openxai,
  title = {{OpenXAI}: {Towards} a {Transparent Evaluation} of {Model Explanations} },
  author = {Agarwal, Chirag and Krishna, Satyapriya and Saxena, Eshika and Pawelczyk, Martin and Johnson, Nari and Puri, Isha and Zitnik, Marinka and Lakkaraju, Himabindu},
  date = {2022-12-06},
  journaltitle = {Advances in Neural Information Processing Systems},
  volume = {35},
  pages = {15784--15799}
}

@article{cortez2013using,
  title = {Using Sensitivity Analysis and Visualization Techniques to Open Black Box Data Mining Models},
  author = {Cortez, Paulo and Embrechts, Mark J.},
  date = {2013-03-10},
  journaltitle = {Information Sciences},
  volume = {225},
  pages = {1--17},
}

@inproceedings{bastings2022will,
  title = {“{Will You Find These Shortcuts}?” {A Protocol} for {Evaluating} the {Faithfulness} of {Input Salience Methods} for {Text Classification} },
  booktitle = {Proceedings of the 2022 {Conference} on {Empirical Methods} in {Natural Language Processing} },
  author = {Bastings, Jasmijn and Ebert, Sebastian and Zablotskaia, Polina and Sandholm, Anders and Filippova, Katja},
  date = {2022-12},
  pages = {976--991},
  location = {Abu Dhabi, United Arab Emirates},
  eventtitle = {{EMNLP} 2022}
}

@inproceedings{lin2014microsoft,
  title={Microsoft coco: Common objects in context},
  author={Lin, Tsung-Yi and Maire, Michael and Belongie, Serge and Hays, James and Perona, Pietro and Ramanan, Deva and Doll{\'a}r, Piotr and Zitnick, C Lawrence},
  booktitle={Computer vision--ECCV 2014: 13th European conference, zurich, Switzerland, September 6-12, 2014, proceedings, part v 13},
  pages={740--755},
  year={2014},
  organization={Springer}
}

@article{zhou2017places,
   title={Places: A 10 million Image Database for Scene Recognition},
   author={Zhou, Bolei and Lapedriza, Agata and Khosla, Aditya and Oliva, Aude and Torralba, Antonio},
   journal={IEEE Transactions on Pattern Analysis and Machine Intelligence},
   year={2017},
   publisher={IEEE}
 }

@inproceedings{alhamoud2025vision,
  title={Vision-language models do not understand negation},
  author={Alhamoud, Kumail and Alshammari, Shaden and Tian, Yonglong and Li, Guohao and Torr, Philip HS and Kim, Yoon and Ghassemi, Marzyeh},
  booktitle={Proceedings of the Computer Vision and Pattern Recognition Conference},
  pages={29612--29622},
  year={2025}
}

@inproceedings{rahmanzadehgervi2024vision,
  title={Vision language models are blind},
  author={Rahmanzadehgervi, Pooyan and Bolton, Logan and Taesiri, Mohammad Reza and Nguyen, Anh Totti},
  booktitle={Proceedings of the Asian Conference on Computer Vision},
  pages={18--34},
  year={2024}
}

@inproceedings{chenwe,
  title={Are We on the Right Way for Evaluating Large Vision-Language Models?},
  author={Chen, Lin and Li, Jinsong and Dong, Xiaoyi and Zhang, Pan and Zang, Yuhang and Chen, Zehui and Duan, Haodong and Wang, Jiaqi and Qiao, Yu and Lin, Dahua and others},
  booktitle={The Thirty-eighth Annual Conference on Neural Information Processing Systems}
}

@inproceedings{yanworse,
  title={Worse than Random? An Embarrassingly Simple Probing Evaluation of Large Multimodal Models in Medical VQA},
  author={Yan, Qianqi and He, Xuehai and Yue, Xiang and Wang, Xin Eric},
  booktitle={GenAI for Health: Potential, Trust and Policy Compliance}
}

@techreport{WahCUB_200_2011,
	Title = {Caltech UCSD Birds 200 2011},
	Author = {Wah, C. and Branson, S. and Welinder, P. and Perona, P. and Belongie, S.},
	Year = {2011},
	Institution = {California Institute of Technology},
	Number = {CNS-TR-2011-001}
}

@article{chen2019looks,
  title={This looks like that: deep learning for interpretable image recognition},
  author={Chen, Chaofan and Li, Oscar and Tao, Daniel and Barnett, Alina and Rudin, Cynthia and Su, Jonathan K},
  journal={Advances in neural information processing systems},
  volume={32},
  year={2019}
}

@inproceedings{nauta2023pip,
  title={Pip-net: Patch-based intuitive prototypes for interpretable image classification},
  author={Nauta, Meike and Schl{\"o}tterer, J{\"o}rg and Van Keulen, Maurice and Seifert, Christin},
  booktitle={Proceedings of the IEEE/CVF Conference on Computer Vision and Pattern Recognition},
  pages={2744--2753},
  year={2023}
}

@article{pares2022mame,
  title={The MAMe dataset: on the relevance of high resolution and variable shape image properties},
  author={Par{\'e}s, Ferran and Arias-Duart, Anna and Garcia-Gasulla, Dario and Campo-Franc{\'e}s, Gema and Viladrich, Nina and Ayguad{\'e}, Eduard and Labarta, Jes{\'u}s},
  journal={Applied Intelligence},
  volume={52},
  number={10},
  pages={11703--11724},
  year={2022},
  publisher={Springer}
}

@inproceedings{quattoni2009recognizing,
  title={Recognizing indoor scenes},
  author={Quattoni, Ariadna and Torralba, Antonio},
  booktitle={2009 IEEE Conference on Computer Vision and Pattern Recognition},
  pages={413--420},
  year={2009},
  organization={IEEE}
}

@article{russakovsky2015imagenet,
  title={Imagenet large scale visual recognition challenge},
  author={Russakovsky, Olga and Deng, Jia and Su, Hao and Krause, Jonathan and Satheesh, Sanjeev and Ma, Sean and Huang, Zhiheng and Karpathy, Andrej and Khosla, Aditya and Bernstein, Michael and others},
  journal={International journal of computer vision},
  volume={115},
  number={3},
  pages={211--252},
  year={2015},
  publisher={Springer}
}

@article{krizhevsky2009learning,
  title={Learning multiple layers of features from tiny images},
  author={Krizhevsky, Alex and Hinton, Geoffrey and others},
  year={2009},
  publisher={Toronto, ON, Canada}
}

@article{rubner2000earth,
  title={The earth mover's distance as a metric for image retrieval},
  author={Rubner, Yossi and Tomasi, Carlo and Guibas, Leonidas J},
  journal={International journal of computer vision},
  volume={40},
  pages={99--121},
  year={2000},
  publisher={Springer}
}

@article{judd2012benchmark,
  title={A benchmark of computational models of saliency to predict human fixations},
  author={Judd, Tilke and Durand, Fr{\'e}do and Torralba, Antonio},
  year={2012}
}

@inproceedings{rao2022better,
  title = {Towards {Better Understanding Attribution Methods} },
  booktitle = {2022 {IEEE}/{CVF Conference} on {Computer Vision} and {Pattern Recognition} ({CVPR})},
  author = {Rao, Sukrut and Bohle, Moritz and Schiele, Bernt},
  date = {2022-06},
  pages = {10213--10222},
  location = {New Orleans, LA, USA},
  eventtitle = {2022 {IEEE}/{CVF Conference} on {Computer Vision} and {Pattern Recognition} ({CVPR})},
  isbn = {978-1-6654-6946-3}
}

@inproceedings{arias2023confusion,
  title={A confusion matrix for evaluating feature attribution methods},
  author={Arias-Duart, Anna and Mariotti, Ettore and Garcia-Gasulla, Dario and Alonso-Moral, Jose Maria},
  booktitle={Proceedings of the IEEE/CVF Conference on Computer Vision and Pattern Recognition},
  pages={3709--3714},
  year={2023}
}

@inproceedings{desai2020ablationcam,
  title = {Ablation-{CAM}: {Visual Explanations} for {Deep Convolutional Network} via {Gradient-free Localization} },
  booktitle = {2020 {IEEE Winter Conference} on {Applications} of {Computer Vision} ({WACV})},
  author = {Desai, Saurabh and Ramaswamy, Harish G.},
  date = {2020-03},
  pages = {972--980},
  location = {Snowmass Village, CO, USA},
  eventtitle = {2020 {IEEE Winter Conference} on {Applications} of {Computer Vision} ({WACV})},
  isbn = {978-1-7281-6553-0}
}

@article{selvaraju2020gradcam,
  title = {Grad-{CAM}: {Visual Explanations} from {Deep Networks} via {Gradient-Based Localization} },
  author = {Selvaraju, Ramprasaath R. and Cogswell, Michael and Das, Abhishek and Vedantam, Ramakrishna and Parikh, Devi and Batra, Dhruv},
  date = {2020-02-11},
  journaltitle = {International Journal of Computer Vision},
  volume = {128},
  number = {2},
  eprint = {1610.02391},
  eprinttype = {arXiv},
  pages = {336--359}
}

@article{springenberg2014striving,
  title = {Striving for {Simplicity}: {The All Convolutional Net} },
  author = {Springenberg, Jost Tobias and Dosovitskiy, Alexey and Brox, Thomas and Riedmiller, Martin},
  date = {2014-12-21},
  journaltitle = {3rd International Conference on Learning Representations, ICLR 2015 - Workshop Track Proceedings},
  eprint = {1412.6806},
  eprinttype = {arXiv},
  pages = {1--14}
}

@inproceedings{shrikumar2019learning,
  title={Learning important features through propagating activation differences},
  author={Shrikumar, Avanti and Greenside, Peyton and Kundaje, Anshul},
  booktitle={International conference on machine learning},
  pages={3145--3153},
  year={2017},
  organization={PMlR}
}

@article{sundararajan2017axiomatic,
  title = {Axiomatic {Attribution} for {Deep Networks} },
  author = {Sundararajan, Mukund and Taly, Ankur and Yan, Qiqi},
  date = {2017},
  journaltitle = {Proceedings of the 34th International Conference on Machine Learning}
}

@inproceedings{brendel2018approximating,
  title = {Approximating {CNNs} with {Bag-of-local-Features} Models Works Surprisingly Well on {ImageNet} },
  booktitle = {International Conference on Learning Representations},
  author = {Brendel, Wieland and Bethge, Matthias},
  date = {2019}
}

@article{bohle2024bcos,
  title = {B-{Cos Alignment} for {Inherently Interpretable CNNs} and {Vision Transformers} },
  author = {Böhle, Moritz and Singh, Navdeeppal and Fritz, Mario and Schiele, Bernt},
  date = {2024-06},
  journaltitle = {IEEE Transactions on Pattern Analysis and Machine Intelligence},
  volume = {46},
  number = {6},
  pages = {4504--4518}
}

@inproceedings{kindermans2018learning,
    title={Learning how to explain neural networks: PatternNet and PatternAttribution},
    author={Pieter-Jan Kindermans and Kristof T. Schütt and Maximilian Alber and Klaus-Robert Müller and Dumitru Erhan and Been Kim and Sven Dähne},
    booktitle={International Conference on Learning Representations},
    year={2018},
    url={https://openreview.net/forum?id=Hkn7CBaTW},
}

@inproceedings{hesse2021fast,
  title = {Fast Axiomatic Attribution for Neural Networks},
  booktitle = {Advances in Neural Information Processing Systems},
  author = {Hesse, Robin and Schaub-Meyer, Simone and Roth, Stefan},
  date = {2021}
}

@inproceedings{chefer2021transformer,
  title = {Transformer {Interpretability Beyond Attention Visualization} },
  author = {Chefer, Hila and Gur, Shir and Wolf, Lior},
  date = {2021},
  pages = {782--791},
  eventtitle = {Proceedings of the {IEEE}/{CVF Conference} on {Computer Vision} and {Pattern Recognition}}
}

@article{zhang2018topdown,
  title = {Top-{Down Neural Attention} by {Excitation Backprop} },
  author = {Zhang, Jianming and Bargal, Sarah Adel and Lin, Zhe and Brandt, Jonathan and Shen, Xiaohui and Sclaroff, Stan},
  date = {2018-10},
  journaltitle = {International Journal of Computer Vision},
  volume = {126},
  number = {10},
  pages = {1084--1102}
}

@article{smilkov2017smoothgrad,
  title = {{SmoothGrad}: {Removing} Noise by Adding Noise},
  author = {Smilkov, Daniel and Thorat, Nikhil and Kim, Been and Viégas, Fernanda and Wattenberg, Martin},
  date = {2017},
  journaltitle = {arXiv},
  eprint = {1706.03825},
  eprinttype = {arXiv}
}

@misc{adebayo2018local,
  title = {Local Explanation Methods for Deep Neural Networks Lack Sensitivity to Parameter Values},
  author = {Adebayo, Julius and Gilmer, Justin and Goodfellow, Ian and Kim, Been},
  date = {2018}
}

@article{jiang2021layercam,
  title = {{LayerCAM}: {Exploring Hierarchical Class Activation Maps} for {Localization} },
  author = {Jiang, Peng-Tao and Zhang, Chang-Bin and Hou, Qibin and Cheng, Ming-Ming and Wei, Yunchao},
  date = {2021},
  journaltitle = {IEEE Transactions on Image Processing},
  volume = {30},
  pages = {5875--5888}
}

@inproceedings{wang2020scorecam,
  title = {Score-{CAM}: {Score-Weighted Visual Explanations} for {Convolutional Neural Networks} },
  booktitle = {2020 {IEEE}/{CVF Conference} on {Computer Vision} and {Pattern Recognition Workshops} ({CVPRW})},
  author = {Wang, Haofan and Wang, Zifan and Du, Mengnan and Yang, Fan and Zhang, Zijian and Ding, Sirui and Mardziel, Piotr and Hu, Xia},
  date = {2020-06},
  pages = {111--119},
  location = {Seattle, WA, USA},
  eventtitle = {2020 {IEEE}/{CVF Conference} on {Computer Vision} and {Pattern Recognition Workshops} ({CVPRW})},
  isbn = {978-1-7281-9360-1}
}

@article{petsiuk2018rise,
  title = {{RISE}: {Randomized Input Sampling} for {Explanation} of {Black-box Models} },
  author = {Petsiuk, Vitali and Das, Abir and Saenko, Kate},
  date = {2018-06-19},
  journaltitle = {British Machine Vision Conference 2018, BMVC 2018},
  volume = {1},
  eprint = {1806.07421},
  eprinttype = {arXiv}
}

@article{muddamsetty2022visual,
  title = {Visual Explanation of Black-Box Model:~{Similarity Difference} and {Uniqueness} ({SIDU}) Method},
  author = {Muddamsetty, Satya M. and Jahromi, Mohammad N. S. and Ciontos, Andreea E. and Fenoy, Laura M. and Moeslund, Thomas B.},
  date = {2022-07-01},
  journaltitle = {Pattern Recognition},
  volume = {127},
  pages = {108604}
}

@article{plumb2018model,
  title = {Model Agnostic Supervised Local Explanations},
  author = {Plumb, Gregory and Molitor, Denali and Talwalkar, Ameet S},
  date = {2018},
  journaltitle = {Advances in neural information processing systems},
  volume = {31}
}

@inproceedings{abnar2020quantifying,
  title = {Quantifying Attention Flow in Transformers},
  booktitle = {Proceedings of the 58th Annual Meeting of the Association for Computational Linguistics},
  author = {Abnar, Samira and Zuidema, Willem},
  date = {2020},
  pages = {4190--4197}
}

@article{lundberg2017unified,
  title = {A Unified Approach to Interpreting Model Predictions},
  author = {Lundberg, Scott M and Lee, Su-In},
  date = {2017},
  journaltitle = {Advances in neural information processing systems},
  volume = {30}
}

@inproceedings{kim2018interpretability,
  title = {Interpretability beyond Feature Attribution: {Quantitative} Testing with Concept Activation Vectors ({TCAV})},
  booktitle = {Proceedings of the 35th International Conference on Machine Learning},
  author = {Kim, Been and Wattenberg, Martin and Gilmer, Justin and Cai, Carrie and Wexler, James and Viegas, Fernanda and {sayres}, Rory},
  date = {2018-07-10/2018-07-15},
  series = {Proceedings of Machine Learning Research},
  volume = {80},
  pages = {2668--2677}
}

@inproceedings{calderon-reichart-2025-behalf,
    title = "On Behalf of the Stakeholders: Trends in {NLP} Model Interpretability in the Era of {LLM}s",
    author = "Calderon, Nitay  and
      Reichart, Roi",
    editor = "Chiruzzo, Luis  and
      Ritter, Alan  and
      Wang, Lu",
    booktitle = "Proceedings of the 2025 Conference of the Nations of the Americas Chapter of the Association for Computational Linguistics: Human Language Technologies (Volume 1: Long Papers)",
    month = apr,
    year = "2025",
    address = "Albuquerque, New Mexico",
    publisher = "Association for Computational Linguistics",
    url = "https://aclanthology.org/2025.naacl-long.29/",
    doi = "10.18653/v1/2025.naacl-long.29",
    pages = "656--693",
    ISBN = "979-8-89176-189-6"
}
